\documentclass[10pt,journal,compsoc]{IEEEtran}

\ifCLASSOPTIONcompsoc
  \usepackage[nocompress]{cite}
\else
  \usepackage{cite}
\fi
\usepackage{graphicx}
\usepackage{amsmath}
\usepackage{amssymb}
\usepackage{booktabs}
\usepackage{bbding}
\usepackage{epsfig}
\usepackage{multirow}
\usepackage{threeparttable}  
\usepackage{color}
\usepackage{array}
\usepackage{makecell}
\usepackage{stfloats}
\usepackage[table]{xcolor}
\usepackage[misc]{ifsym}
\usepackage[accsupp]{axessibility}
\usepackage{hyperref}
\usepackage{subfigure}
\usepackage{bbm}

\ifCLASSINFOpdf
\else
\fi

\begin{document}

\title{SAM 2++: Tracking Anything at Any Granularity}

\author{
Jiaming~Zhang\IEEEauthorrefmark{2}, Cheng~Liang\IEEEauthorrefmark{2}, Yichun~Yang\IEEEauthorrefmark{2}, Chenkai~Zeng\IEEEauthorrefmark{2}, Yutao~Cui, Xinwen~Zhang, Xin~Zhou, Kai~Ma, Gangshan~Wu,~\IEEEmembership{Member,~IEEE} and~Limin~Wang,~\IEEEmembership{Senior Member,~IEEE}
\IEEEcompsocitemizethanks{\IEEEcompsocthanksitem \IEEEauthorrefmark{2} Jiaming Zhang, Cheng Liang, Yichun Yang, and Chenkai Zeng contributed equally.
\IEEEcompsocthanksitem Jiaming Zhang, Cheng Liang, Yichun Yang, Chenkai Zeng, Yutao Cui, Xinwen Zhang, Xin Zhou, and Gangshan Wu are with the State Key Laboratory for Novel Software Technology, Nanjing University, Nanjing 210023, China (e-mail: jiamming.zhang@gmail.com; lc1190301804@outlook.com; yichunyang@smail.nju.edu.cn; chenkai.zeng@smail.nju.edu.cn; cuiyutao@smail.nju.edu.cn; 502024370064@smail.nju.edu.cn; 522024330134@smail.nju.edu.cn; gswu@nju.edu.cn).
\IEEEcompsocthanksitem Kai Ma is with Tencent, Shenzhen 518054, China (e-mail: kylekma@tencent.com).
\IEEEcompsocthanksitem Limin Wang is with the State Key Laboratory for Novel Software Technology, Nanjing University, Nanjing 210023, China, and also with the Shanghai Artificial Intelligence Laboratory, Shanghai 200232, China (e-mail: lmwang@nju.edu.cn).
}
}


\IEEEtitleabstractindextext{%
\begin{abstract}
Video tracking aims at finding the specific target in subsequent frames given its diverse initial states.
Due to the varying granularity of target states across different tasks, most existing trackers are tailored to a single task and rely heavily on custom-designed modules. This specificity limits their generalization, preventing them from effectively utilizing multi-task training data and leading to redundancy in both model design and parameters.
Although recent unified vision models share partial architectures across tasks, they usually retain task-specific interfaces and overlook the common tracking principle behind different granularities, leaving a gap for truly unified video tracking.
To unify video tracking tasks, we present SAM 2++, a unified framework that can handle target states at different granularities, including masks, boxes, and points, through an integrated design of prompt encoding, output decoding, and memory representation.
First, to handle different target granularities, we design task-specific prompts that map diverse task inputs into general prompt embeddings, together with a Unified Decoder that produces task results in a common output form without redesigning the overall pipeline.
Next, to satisfy memory matching, the core operation of tracking, we introduce a task-adaptive memory mechanism that unifies memory across different granularities while preserving their distinct state semantics, preventing full parameter sharing from causing interference across granularities.
Finally, we introduce Tracking-Any-Granularity, the first large and diverse video tracking dataset with rich annotations at three granularities.
It is constructed through a customized data engine with phased manual annotation and model-assisted completion, providing a comprehensive resource for training, benchmarking, and analyzing unified tracking models.
With the help of our Tracking-Any-Granularity dataset, comprehensive experiments on multiple benchmarks confirm that SAM 2++ sets a new state of the art across diverse tracking tasks at different granularities, establishing a unified and robust tracking framework.
Code, trained models, and dataset are available at \url{https://github.com/MCG-NJU/SAM2-Plus}.
\end{abstract}

}

\maketitle

\IEEEdisplaynontitleabstractindextext

\IEEEpeerreviewmaketitle

\ifCLASSOPTIONcompsoc
\IEEEraisesectionheading{\section{Introduction}\label{sec:introduction}}
\else
\section{Introduction}
\label{sec:introduction}
\fi

\IEEEPARstart{V}{ideo} tracking has been a fundamental task in computer vision for decades, aiming to estimate the state of an arbitrary target in video
sequences given its initial status.
Despite sharing this core objective, the tracking domain has fragmented into several independent sub-tasks based on different target granularities, including Single Object Tracking~\cite{lasot,TrackingNet,got10k} (SOT) with bounding box, Video Object Segmentation~\cite{davis17,youtube-vos,LVOS_V1} (VOS) with precise pixel-level mask, and Point Tracking~\cite{badja,zheng2023pointodyssey,tapvid} with tiny points.
All three tasks follow user-specified targets over time, but represent target states as coarse object locations, dense object shapes, or fine-grained keypoints at different granularities.
This fragmentation based on state granularity has led most video tracking research to focus on a specific task and propose specialized designs only for that task.
While this design trend enhances tracking performance, it limits the generalization ability of tracking and training across multiple tasks and results in redundancy in both model design and parameters.
To unify tasks, current unified vision models typically share feature extraction backbones while employing task-specific branches~\cite{zhu2022uniperceiver}, convert those tasks into a seq2seq framework~\cite{chen2021pix2seq}, or share one appearance model for either propagation or association~\cite{wang2021UniTrack, yu2024unifiedtt, Unicorn, UNINEXT, OmniTracker}.
Still, such unification often remains at shared components rather than a common representation of tracking states.
However, they choose to provide different interfaces for different tasks, rather than seeking a unified visual representation of tracking targets, and ignore the point tracking task.

Unlike them, we observe that these seemingly disparate tracking paradigms fundamentally differ primarily in their state granularity, while sharing the \textit{memory matching} strategy, where the model encodes the previous state into memory and matches the current features with the stored memory when a new frame is received.
This shared process offers a natural entry point for bridging different target representations.
Based on this strategy, we decide to unify target states at three different granularities through a uniform memory representation.
Recently, Segment Anything Model 2~\cite{ravi2024sam2}, a strong foundational model, has been proposed for high-quality video object segmentation given various prompts.
Due to its flexible prompt mechanism and powerful mask tracking capabilities, we extend this model to track arbitrary granularity, termed as \textbf{SAM 2++}.

\begin{figure*}[t]
    \centering
    \includegraphics[width=\linewidth]{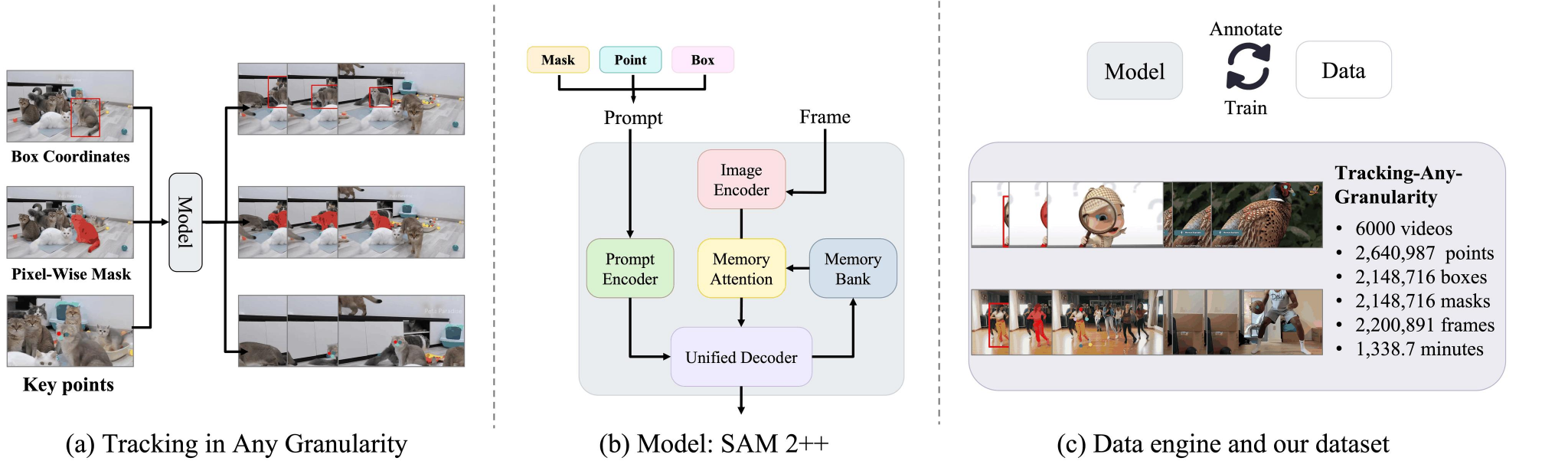}
    \caption{The overall framework of SAM 2++, including (a) tracking at any granularity task, (b) our unified tracking framework, and (c) our Tracking-Any-Granularity dataset collected through our data engine. SAM 2++ is capable of tracking at any granularity.}
    \label{fig:overview}
\end{figure*}

Our work includes a model and a dataset (see Fig.~\ref{fig:overview}).
To ensure generalized tracking at different granularities, we start by designing \textit{task-specific prompts} and a \textit{unified decoder}.
Specifically, we introduce task-specific prompts to encode different task inputs into general prompt embeddings.
As for the diverse task output, our unified decoder, which is extended from the Mask Decoder of SAM 2, unifies diverse task results into a unified form pre-output.
With this design, task inputs remain compatible with the SAM 2 prompt interface, while task outputs are converted into representations for the shared tracking pipeline.
Next, we found that full parameter sharing in task-mixed training leads to performance degradation on all tasks, due to the different memory requirements of different tracking tasks.
To address this, we introduce a \textit{task-adaptive memory mechanism}, which adjusts memory representations in response to the unique requirements of each task.
This mechanism not only helps to offset the adverse effects of full parameter sharing on the memory mechanism but also achieves mutual promotion among multiple tasks.
Instead of forcing all tasks to write identical memory features, the task-adaptive memory keeps the shared tracking process while allowing each granularity to maintain its own state semantics.

To support tracking at any granularity in videos, we further construct a large and diverse video tracking dataset, termed \textbf{T}racking-\textbf{A}ny-\textbf{G}ranularity (TAG), through a data engine.
The data engine produces training data through an interactive process, where annotators manually label data at varying intervals in different phases. Then, after phased training, the model is used to annotate the remaining frames, achieving efficient and accurate expansion of the dataset.
Unlike most existing video tracking datasets, our dataset provides high-quality annotations at three granularities, including \textit{segmentation masks}, \textit{bounding boxes}, and \textit{key points}, resulting in a vital resource for training and benchmarking unified tracking models.
These annotations make TAG suitable for both mixed training and cross-granularity evaluation.
Extensive experiments on several benchmarks across tasks demonstrate that SAM 2++ enables tracking at any granularity with a unified model architecture and consistently outperforms task-specific models in all three tasks.
The main contributions are summarized as follows:
\begin{itemize}
    \item We propose a unified framework, termed SAM 2++, towards tracking targets at any granularity by task-specific prompts, a unified decoder, and a task-adaptive memory mechanism for various granularities, with a shared memory representation.
    \item We build a data engine that produces training data through an interactive process both manually and automatically, resulting in a new large-scale object tracking dataset, \textbf{T}racking-\textbf{A}ny-\textbf{G}ranularity (TAG), with high-quality annotations in various granularities for unified training and evaluation.
    \item Experiments show that SAM 2++ enables accurate tracking at various granularities, consistently surpassing the performance of task-specific models.
\end{itemize}

\section{Related work}
\textbf{Segment Anything Model.}
SAM~\cite{kirillov2023seganysam1} is a foundational model for high-quality segmentation given various prompts, and SAM 2~\cite{ravi2024sam2} extends it to video with streaming memory, effectively handling motion and occlusion, and they inspire many variant models.
In the image domain, HQ-SAM~\cite{sam_hq} and SAMRefiner~\cite{lin2025samrefiner} improve mask quality and fine-grained details, CAT-SAM~\cite{xiao2024cat} and SAM-Adapter~\cite{chen2023samadapter} adapt SAM to specialized domains, while FastSAM~\cite{zhao2023fastsam}, MobileSAM~\cite{zhang2023mobilesam}, and EfficientSAM~\cite{xiong2024efficientsam} reduce its computational cost.
Grounded SAM~\cite{ren2024groundedSAM} and Semantic-SAM~\cite{li2024semanticsam} extend SAM-style segmentation toward open-world and granularity-aware perception.
In the video domain, SAM2Long~\cite{ding2024sam2long} employs constrained tree search to reduce error accumulation, SAMURAI~\cite{yang2024samurai} uses the Kalman filter to select motion-aware memory, while DAM4SAM~\cite{videnovic2024distractor} introduces a distractor-aware memory.
SAM-PT~\cite{rajic2025sampt} combines SAM with point tracking for point-centric video segmentation, EfficientTAM~\cite{xiong2025efficienttam} and EdgeTAM~\cite{zhou2025edgetam} improve the efficiency of SAM 2-style track-anything models, and Segment Any Motion~\cite{huang2025segmentanymotion} explores motion-aware video segmentation.
SAMWISE~\cite{cuttano2024samwise} and AL-Ref-SAM-2~\cite{huang2025unleashing} add additional prompts for more referring tasks.
Despite these innovations addressing specific challenges across the segmentation and tracking spectrum, they remain largely isolated solutions optimized for individual tasks, overlooking potential synergies between related vision problems and requiring separate specialized implementations for each application domain.

\textbf{Unified Vision Models.}
Recent years have witnessed significant progress in developing unified vision models that handle multiple tasks through shared architectures and demonstrate strong generalizability and flexibility.
Pix2Seq~\cite{chen2021pix2seq,chen2022piexl2seq2} reformulates vision tasks as sequence generation problems, while Uni-Perceiver~\cite{zhu2022uniperceiver,li2023uniperceiver2} establishes unified representation spaces across modalities with shared encoders and decoders.
Unified-IO~\cite{DBLP:conf/iclr/LuCZMK23} and Painter~\cite{DBLP:conf/cvpr/WangWCS023} further explore generalist models with unified task interfaces, and SegGPT~\cite{DBLP:conf/iccv/WangZCWS023} studies in-context segmentation across different segmentation forms.
UniTrack~\cite{wang2021UniTrack} demonstrates that video tracking tasks can be solved by a single appearance model with task-specific heads, while Unicorn~\cite{Unicorn}, UNINEXT~\cite{UNINEXT}, MITS~\cite{MITS}, and OmniTracker~\cite{OmniTracker} unify different object-level tracking or segmentation paradigms.
Despite their impressive capabilities, these unified approaches predominantly focus on object-level tasks while neglecting finer-grained tasks such as point tracking.
Furthermore, most require joint training from scratch on multiple large-scale datasets, demanding substantial computational resources that limit their accessibility and development.
Moreover, they do not take into account unifying tracking tasks with various granularities through a unified representation.

\textbf{Tracking Datasets}
Existing video tracking datasets are usually constructed for a specific target granularity.
VOS datasets such as DAVIS-2017~\cite{davis17}, YoutubeVOS~\cite{youtube-vos}, LVOS~\cite{LVOS_V1,LVOS_V2}, and MOSE~\cite{MOSE} mainly provide mask annotations, with motivations ranging from precise labels and large-scale coverage to long-term tracking and complex scenarios.
SOT datasets such as LaSOT~\cite{lasot}, GOT-10k~\cite{got10k}, TrackingNet~\cite{TrackingNet}, and others focus on box trajectories for object-level localization, covering challenges such as large-scale training, long-term tracking, high-frame-rate videos, UAV scenes, language-guided tracking, and abundant object categories.
Point tracking datasets such as Perception Test~\cite{patraucean2023perception}, PointOdyssey~\cite{zheng2023pointodyssey}, and TAP-Vid~\cite{tapvid} provide keypoint or arbitrary-point trajectories for fine-grained evaluation, including multi-modal, long-term, real-world, and arbitrary-point settings.
Although these datasets have advanced their respective tasks, they are typically annotated for a single task and a single target-state granularity.
In contrast, our Tracking-Any-Granularity (TAG) dataset provides segmentation masks, bounding boxes, and key points for the same video tracking data, making it the only dataset in our comparison that supports all three granularities simultaneously for unified tracking training and evaluation.

\section{Preliminaries: Segment Anything Model}
The Segment Anything Model (SAM)~\cite{kirillov2023seganysam1} is a milestone vision foundation model for class-agnostic image segmentation. It flexibly handles various prompts (box, point, mask) by encoding them into a unified embedding and has established an iterative data engine with model-assisted labeling to address dataset limitations.
SAM 2~\cite{ravi2024sam2} extends SAM to promptable video segmentation by introducing a streaming memory that stores previous target information and predictions. It comprises four main components:
(i) a hierarchical image encoder that encodes each frame ${I}_{img}$ into image embeddings $F_{img}$,
(ii) a prompt encoder,
(iii) a memory mechanism (memory encoder, memory bank, memory attention),
and (iv) a mask decoder for prediction.

\textbf{Prompt Encoder.}
SAM 2 follows the prompt encoder design from SAM to support three types of user inputs, including positive/negative points, bounding boxes, and masks.
The point prompt ${I}_{point} \in \mathbb{R}^{N_{point} \times 2}$ and box prompt ${I}_{box} \in \mathbb{R}^{2 \times 2}$ (seen as two corner points) can be represented as sparse embeddings ${\mathcal{P}_{sparse} \in \mathbb{R}^{N_{point} \times C}}$ by their point location and learnable embedding parameters $\varepsilon_{sparse}^{point}, \varepsilon_{sparse}^{box}$ which encodes the type of each point.
As for the mask prompt ${I}_{mask} \in \mathbb{R}^{1 \times H \times W}$, the model adopts convolutions to map and downscale them as dense embedding ${\mathcal{P}_{dense} \in \mathbb{R}^{C \times H/16 \times W/16}}$.
In summary, the processing of Prompt Encoder can be written as:
\begin{equation}
\begin{aligned}
    \mathcal{P}_{sparse} &= [\textit{PE}({I}_{point}) + \varepsilon_{sparse}^{point}; \textit{PE}({I}_{box}) + \varepsilon_{sparse}^{box}], \\
    \mathcal{P}_{dense} &= \mathbf{Conv}_{dense}({I}_{mask}), \\
\end{aligned}
\end{equation}
where the $\text{PE}$ represents positional encoding operation.

\textbf{Mask Decoder.}
The mask decoder takes prompt embedding $\mathcal{P}_{sparse}$ and $\mathcal{P}_{dense}$, memory-conditioned image embeddings $\bar{F}_{img} \in \mathbb{R}^{C/4 \times H/16 \times W/16}$ (which we will explain later), and a set of learnable tokens $\mathcal{E}_{tokens}$ as inputs.
The learnable tokens contain an existence token $\varepsilon_{obj} \in \mathbb{R}^{C}$ to predict whether the target exists, an IoU token $\varepsilon_{iou} \in \mathbb{R}^{C}$ to predict the result accuracy, and multiple mask tokens $\varepsilon_{mask}^{N} \in \mathbb{R}^{N \times C}$ used to obtain $N$ mask candidates.
To fuse the prompt embedding, a Two-Way Transformer $\mathbf{twTrans}$ \cite{ravi2024sam2} processes them as:
\begin{equation}
\begin{aligned}
\tilde{F}_{img}, [\tilde{\mathcal{P}}_{sparse}; \tilde{\mathcal{E}}_{tokens}] = \mathbf{twTrans}(  \\
\bar{F}_{img} + \mathcal{P}_{dense}, [\mathcal{P}_{sparse}; \mathcal{E}_{tokens}]).
\end{aligned}
\end{equation}

After that, the output token embeddings $\tilde{\mathcal{E}}_{tokens}$ are split into $\tilde{\varepsilon}_{obj}$ for predicting existence $O_{obj}$, $\tilde{\varepsilon}_{iou}$ for producing IoU scores $O_{IoU}^{N}$, and $\tilde{\varepsilon}_{mask}^{N}$ for generating mask output as:
\begin{equation}
M^{i}_{mask} = \text{Interpolate}(\tilde{F}_{img} \cdot \tilde{\varepsilon}_{mask}^{i}),
\end{equation}
where the $M^{i}_{mask}$ represents the $i_{th}$ candidate mask prediction rated by corresponding iou score.

\textbf{Memory.}
The memory encoder $\mathbf{MemEn}$ processes image embedding $F_{img}$ and the mask prediction $M_{mask}^{*}$ with the highest IoU score to generate memory embedding $\bar{\bar{F}}_{img}$ for the processed frame.
In addition, it introduces object pointer $\varepsilon_{pointer} \in \mathbb{R}^{C}$, which is transformed from the mask token $\tilde{\varepsilon}_{mask}^{*}$, to provide high-level semantic information.
After that, these two kinds of memory are appended to Memory Bank $\mathcal{MB}$ in FIFO mode.
To enable the current frame to obtain past target information, the image embeddings ${F}_{img}$ are not directly fed to the Mask Decoder, but instead conditioned on memories from Memory Bank as $\bar{F}_{img}$ by cross-attention in Memory Attention $\mathbf{MemAttn}$.

\section{SAM 2++: Unified Tracking Model}
\begin{figure*}[t]
    \centering
    \includegraphics[width=\linewidth]{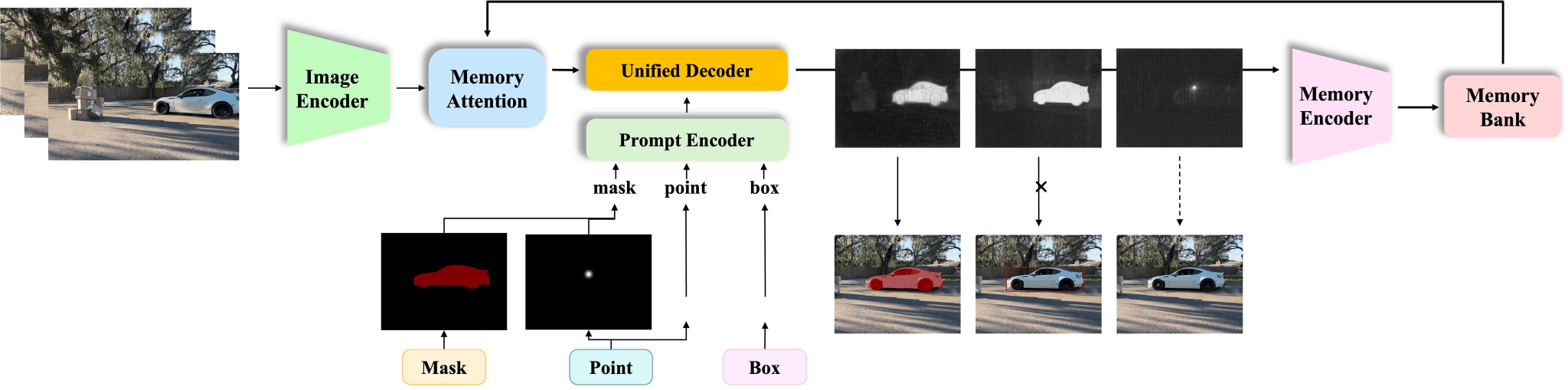}
    \vspace{-6mm}
    \caption{The SAM 2++ architecture. When a new frame is received, the result is conditioned on the new prompt \textit{and/or} stored memories. The initial target state at any granularity is converted into task-specific prompts for unified input. The Unified decoder predicts the task result for the current frame in unified mask form. Finally, the task-adaptive memory transforms diverse target states into unified memory.}
    \label{fig:model}
    \vspace{-4mm}
\end{figure*}

In this section, we present our unified video tracking framework, termed as \textbf{SAM 2 ++}, which extends the SAM 2 model to track any targets in videos at any granularity, including masks, bounding boxes, and points, and the overall pipeline is depicted in Fig.~\ref{fig:model}.
Due to the various task granularities, we introduce \textit{task-specific prompts} to unify task input in different granularities and the \textit{Unified Decoder} to unify diverse task results into a unified form pre-output.
Next, we found that a fully parameter-shared model training results in performance degradation due to the diverse memory requirements across tasks.
To address this, we introduce a \textit{task-adaptive memory mechanism} that dynamically adjusts memory representations according to each task's demand, enhancing the multi-task processing capability.

\subsection{Unified Task Input and Output Processing}

\textbf{Input Unification via Task-Specific Prompt.}
Due to the input of the three tracking tasks having inconsistent granularity, we first unify inputs with task-specific prompts for different tasks, while keeping the original Prompt Encoder unchanged.
This preserves the prompt space of SAM 2 and avoids introducing extra input encoders for individual tracking tasks.
Mask tracking adopts the \{0, 1\} mask ${I}_{mask}$ as the prompt to accurately describe the target shape and boundary, while box tracking uses the bounding box ${I}_{box}$ as the prompt, represented by its top-left and bottom-right corners.
Point tracking uses the point coordinate ${I}_{point}$ together with an additional Gaussian mask ${G}_{point}$, which is centred on the point and parameterised by sigma $\sigma$ and radius $r$ as: $G_{point} = \exp\left(-\frac{\|p - p_0\|^2}{2\sigma^{2}}\right) \cdot \mathbf{1}_{\{\|p - p_0\| \leq r\}}$.
This Gaussian mask highlights the point in mask form, maintains consistency with the output from Unified Decoder and the source for Memory Encoder, and better represents the target than a naive \{0, 1\} mask.
More importantly, we gradually decrease the radius and sigma during training to facilitate smoother convergence and more stable learning.

\textbf{Output Unification via Unified Decoder.}
To unify the output of various tasks, we extend the Mask Decoder of SAM 2 as \textit{Unified Decoder}, which also processes memory-conditioned image embeddings, prompt embeddings, and learnable tokens, and make lightweight task-specific changes on top of the shared mask-form output.
For the SOT task, directly deriving the outer box from the mask output $M^{N}_{box}$ cannot provide an ideal task output, because the complexity of the mask reduces box accuracy and the outer-box operation lacks a direct training gradient.
Instead, we add a Corner-based Head~\cite{yan2021learning_stark}, $\mathbf{CornHead}$, which explicitly optimizes the position and scale of the bounding box and is widely used in the SOT task, to produce box predictions.
As for the PT task, rather than adding an extra coordinate regression head, we obtain point coordinates from the mask output through a soft-argmax operation during training to ensure gradient flow and an argmax operation during inference.
This design aligns point prediction with the Gaussian prompt and the mask source used by Memory Encoder, thereby keeping the point output consistent with the unified memory representation and helping optimize model training.
In summary, our unified decoder can be written as:
\begin{equation}
\begin{aligned}
    M_{mask}^{i} &= \text{Interpolate}(\tilde{F}_{img} \cdot \tilde{\varepsilon}_{mask}^{i}), \\ 
    B_{box}^{i} &= \mathbf{CornHead}(\tilde{F}_{img}, \tilde{\varepsilon}_{mask}^{i}), \\ 
    P_{point}^{i} &= \text{argmax}(\text{Interpolate}(\tilde{F}_{img} \cdot \tilde{\varepsilon}_{mask}^{i})), \\
\end{aligned}
\label{eq:unified_decoder}
\end{equation}
where $M_{mask}^{i}$, $B_{box}^{i}$, and $P_{point}^{i}$ are predictions. represent the $i_{th}$ candidate prediction for three tasks, which are rated by their corresponding iou scores $O_{IoU}^{i}$.

\subsection{Task-adaptive Memory}
Tracking models fundamentally localize targets according to their past states, which requires efficient storage and retrieval ability using a memory-matching paradigm: the model first encodes the previous states into \textit{memory}, then \textit{matches} current features with memory to accurately represent the target when processing a new frame.
Following this paradigm, our model converts mask outputs into memory with Memory Encoder, then applies cross-attention in Memory Attention to match the current frame feature with the feature stored in the memory bank as
\begin{equation}
\begin{aligned}
    \bar{\bar{F}}_{img,\rho} &= \mathbf{MemEn}(F_{img}, M_{\rho}^{*}), \\
    \mathcal{MB}_{\rho} &= \text{FIFO}([\varepsilon_{pointer},\bar{\bar{F}}_{img,\rho}]), \\
     \bar{F}_{img} &= \mathbf{MemAttn}(\mathcal{F}_{img}, \mathcal{MB_{\rho}}, \mathcal{MB_{\rho}}),
\end{aligned}
\label{eq:unified_memory}
\end{equation}
where $\rho$ represents different granularities in various tasks.
However, the mask outputs $M_{\rho}^{*}$ from the three tasks differ in their requirements: in mask tracking, the mask is a precise segmentation; in box tracking, the mask provides coarse localization to assist the box head; in point tracking, the mask is the Gaussian form of the target point.
Based on the above analysis, if the model adopts a full parameter-shared memory module for encoding these diverse mask outputs, it fails to generate task-adaptive memory representations accurately, resulting in memory features failing to meet any requirements of the three tasks.

Therefore, we propose a \textit{task-adaptive memory mechanism}, which relaxes the uniformity by decoupling only the memory components.
Specifically, for the convolutional Memory Encoder, each task has its own copy; for the transformer-based Memory Attention, each task is equipped with an independent LoRA~\cite{DBLP:conf/iclr/HuSWALWWC22/lora}.
This decoupled design effectively meets the diverse needs of different tasks and avoids the performance drop seen in a fully parameter-shared model, while keeping the overall structure consistent and introducing only a minimal parameter increase.
Notably, experiments show that this design enables multiple tasks to promote each other.
Ultimately, we achieve unified multi-task processing with a largely shared memory representation.

\subsection{Training and Inference Details}
\textbf{Training.}
The SAM 2++ training is conducted on 16 H800 GPUs. 
We expand SAM 2++ to three tasks: semi-supervised video object segmentation (mask), single object tracking (box), and online point tracking (point), by processing input, prompt, memory, and output into a unified format used by SAM 2.
Our training process is based on the mask tracking task in SAM 2 and we make minimal modifications to it while adding task-specific requirements from the other tasks. Table \ref{tab:app:train} describes the training settings for three tasks in detail, and other settings not mentioned follow SAM 2.
We initialize SAM 2++ from SAM 2 base and decouple the memory-related modules, including creating a separate copy of the memory encoder and implementing dedicated LoRA parameters for memory attention for each task.

\begin{table*}[t]
    \caption{Hyperparameters and details of SAM 2++ training in three tasks.}
    \label{tab:app:train}
    \centering
    \renewcommand{\arraystretch}{1.12}
    \resizebox{1.0\textwidth}{!}
    {
    \begin{tabular}{cl|ccc}
\toprule
\multicolumn{2}{l|}{\textbf{Settings}} & \multicolumn{1}{c|}{\textbf{mask}}  & \multicolumn{1}{c|}{\textbf{box}} & \textbf{point}  \\
\midrule
\multicolumn{2}{l|}{dataset} & \multicolumn{1}{c|}{\begin{tabular}[c]{@{}c@{}}Tracking-Any-Granularity\\ DAVIS 2017, MOSE, YoutubeVOS 2019\end{tabular}} & \multicolumn{1}{c|}{\begin{tabular}[c]{@{}c@{}}Tracking-Any-Granularity\\ LaSOT, COCO, GOT-10K, TrackingNet\end{tabular}} & \begin{tabular}[c]{@{}c@{}}Tracking-Any-Granularity\\ TAP-Vid Kinetics, PerceptionTest, PointOdyssey\end{tabular}   \\
\midrule
\multicolumn{2}{l|}{sample prob} & \multicolumn{1}{c|}{0.1}   & \multicolumn{1}{c|}{0.4} & 0.5  \\
\midrule
\multicolumn{2}{l|}{shared optimization settings} & \multicolumn{3}{c}{\begin{tabular}[c]{@{}c@{}}batch size: 16; drop path: 0.2 (B+); epochs: 150; resolution: 1024; precision: bfloat16; optimizer: AdamW; $\beta_{1}$, $\beta_{2}$: 0.9, 0.999; \\ gradient clipping: L2, max 0.1; weight decay: 0.1; lr: backbone 5.0e-6, other 3.0e-06; schedule: cosine; layer-wise decay: 0.9 (B+)\end{tabular}} \\
\midrule
\multicolumn{2}{l|}{video augmentation} & \multicolumn{3}{c}{\begin{tabular}[c]{@{}c@{}}hflip; affine (deg: 25, shear: 20) in mask, w/o affine in box and point; resize to 1024 (square);\\ colorjitter (b: 0.1, c: 0.03, s: 0.03, h: null); grayscale (0.05); per-frame colorjitter (b: 0.1, c: 0.05, s: 0.05, h: null)\end{tabular}}   \\
\midrule
\multirow{4}{*}{\textbf{Loss}} & mask losses & \multicolumn{1}{c|}{focal (20), dice (1)} & \multicolumn{1}{c|}{focal (0.5), dice (0.1)}  & focal (20), dice (1)   \\
 & task-specific losses & \multicolumn{1}{c|}{-}  & \multicolumn{1}{c|}{ciou loss (1), box IoU L1  loss   (1)}  & point distance L1  loss (20) \\
 & Iou loss  & \multicolumn{1}{c|}{l1 loss (1)} & \multicolumn{1}{c|}{l1 loss (1)} & l1 loss (1) \\
 & occlusion loss & \multicolumn{1}{c|}{cross-entropy (1)} & \multicolumn{1}{c|}{cross-entropy (1)} & cross-entropy (1)   \\
\midrule
\multicolumn{2}{l|}{input prompt} & \multicolumn{1}{c|}{mask (0.5), noisy box (0.25), sampled points (0.25)} & \multicolumn{1}{c|}{box (0.5), noisy box (0.25), sampled points (0.25)} & point coordinate \& Gaussian mask (1) \\
\midrule
\multicolumn{2}{l|}{\# max. object per frame} & \multicolumn{1}{c|}{3}  & \multicolumn{1}{c|}{1} & 3  \\
\multicolumn{2}{l|}{\# training frames} & \multicolumn{1}{c|}{8}  & \multicolumn{1}{c|}{8} & 8  \\
\multicolumn{2}{l|}{\# init cond frames (w. 0$_{\text{th}}$)} & \multicolumn{1}{c|}{ 1$\sim$2} & \multicolumn{1}{c|}{ 1$\sim$2}  & 1  \\
\multicolumn{2}{l|}{\# corrective frames (w. 0$_{\text{th}}$)} & \multicolumn{1}{c|}{ 1$\sim$2} & \multicolumn{1}{c|}{ 1$\sim$2}  & -  \\
\multicolumn{2}{l|}{\# corrective points} & \multicolumn{1}{c|}{7}  & \multicolumn{1}{c|}{7} & -  \\
\multicolumn{2}{l|}{\# num\_maskmem} & \multicolumn{1}{c|}{7}  & \multicolumn{1}{c|}{7} & 7  \\
\midrule
\multicolumn{2}{l|}{Task-specific Modification} & \multicolumn{3}{c}{Decoupled \& LoRA Memory Attention; Decoupled Memory Encoder; Shared \& LoRA Image Encoder}   \\
\midrule
\multicolumn{2}{l|}{Other Setting} & \multicolumn{1}{c|}{-}  & \multicolumn{1}{c|}{Corner Head} & $\left\{\begin{array}{l}
\text{Ep }0\sim20:\ \text{radius}=50,\ \text{sigma}=16\\
\text{Ep }20\sim50:\ \text{radius}=20,\ \text{sigma}=8\\
\text{Ep }50\sim100:\ \text{radius}=5,\ \text{sigma}=2
\end{array}\right.$ \\
\bottomrule
\end{tabular}

    }
    \vspace{-4mm}
\end{table*}

Training is performed jointly on data of the three tasks.
In addition to our Tracking-Any-Granularity dataset, we used DAVIS-17~\cite{davis17}, YoutubeVOS-19~\cite{youtube-vos} and MOSE~\cite{MOSE} for the mask task, LaSOT~\cite{lasot}, GOT-10k~\cite{got10k}, TrackingNet~\cite{TrackingNet} and COCO~\cite{coco} for the box task, and TAP-Vid Kinetics~\cite{tapvid}, PointOdyssey~\cite{zheng2023pointodyssey}, and PerceptionTest~\cite{patraucean2023perception} for the point task.
To enable the model to be simultaneously capable of all three tasks and to optimize training efficiency, we adopt the strategy of alternating between the three tasks. Specifically, we implement parallelisation by sampling a whole batch at each step of training, which is entirely derived from the data of a particular task. The \textit{sampling probability} is set to 1:4:5 to balance the performance of the three tasks.

We sample 8 frames from each video as a training sequence, randomly choose up to 3 targets (or 1 target box in box tracking) from the objects of this video, and ensure that these sampled targets are visible in the first frame of the sequence.
We randomly select up to 2 frames from the sequence, including the first frame, as \textit{conditional frames} to give these frames initial prompts.
Since we prefer to maintain the \textit{interactive capabilities} of SAM 2, we keep \textit{the interactive prompts} in mask tracking and box tracking during training.
Specifically, we start by deciding whether the conditional frames accept normal or interactive input in this training step with 50\% probability: for normal input, we use ground-truth as initial prompts; for interactive input, we use a noisy bounding box or a positive click from the ground-truth with 50\%-50\% probability.
Alternatively, suppose we use the normal input prompts in conditional frames. In that case, we directly convert them into memory instead of prediction and do not supervise their predictions for this input.
However, the point tracking task requires precise inputs, so we can only provide GT points in the first frame instead of various formats of prompts like the other two tasks.
As for the multi-prediction scenario, when a frame receives no prompt, or at most 1 point (the box prompt can be seen as 2 points), the model will output 3 task predictions and their iou predictions for that frame.

In addition, if interactive input is used as initial prompt, we select up to 2 frames as \textit{corrective frames} to add \textit{corrective clicks} on them: after predicting the selected frame, we sample a positive point from the false positive region between the prediction and ground truth or a negative point from the false negative region as a corrective point, and use it as additional prompt to get a new prediction along with all previous cumulative prompt from that frame. This operation is repeated until 7 corrective points have been added.
In addition, if the box tracking task uses the box format to compute the regional differences between the prediction and the GT, there is an overwhelming problem that the sampled corrective clicks may fall at the boundaries of the box instead of inside the target, which is contrary to the actual interaction. Therefore, we choose to compute the difference in mask format, and use SAM 2 and sam-hq~\cite{sam_hq} with box annotations to obtain the pseudo-GT mask on SOT datasets because of its good segmentation ability.

\textbf{Losses and optimization.}
Following the mask tracking task in SAM 2, we adopt the linear combination of focal loss $\mathcal{L}_{mask}^{focal}$ and dice loss $\mathcal{L}_{mask}^{dice}$ for the mask prediction, L1 loss for the IoU prediction $\mathcal{L}_{L1}^{IoU}$, and cross-entropy loss for object occlusion prediction $\mathcal{L}_{CE}^{obj}$.
During the box tracking task, we adopt the corner head to predict the bounding boxes and add additional ciou loss~\cite{zheng2020diouciou} and L1 loss to supervise the box prediction.
As for the point tracking task, we select the highest probability position from the mask prediction as point prediction, and use soft argmax~\cite{softargmax} during training for making the process derivable instead of the undifferentiable argmax function. Beyond the loss on mask in the form of Gaussian map, we add an L1 loss between the prediction and ground-truth point to directly optimize the distance and accuracy of the points.
For multi-prediction scenario, we only supervise the task predictions (masks, boxes, and points) with the lowest loss, which is a combination of $\mathcal{L}^{mask}$, $\mathcal{L}^{box}$ and $\mathcal{L}^{point}$, but supervise the IoU predictions of all task predictions to learn to synchronise the quality of predictions.
Furthermore, if the target is missing in some frames due to disappearance or cropping, we do not supervise the task predictions or iou predictions on them in all three tasks, but always supervise the occlusion prediction from an MLP head, no matter if the ground-truth exists or not.
In summary, the supervision losses for the three tasks can be written as
\begin{equation}
\begin{aligned}
    \mathcal{L}_{Mask} &= \mathcal{L}_{mask} + \mathcal{L}_{IoU} + \mathcal{L}_{obj} \\
    &= \left [ \lambda_{mask}^{focal} \mathcal{L}_{mask}^{focal} + \lambda_{mask}^{dice} \mathcal{L}_{mask}^{dice}  \right ] \times \mathbbm{1}_{obj} \\ &+ \lambda_{IoU}^{L1}\mathcal{L}_{IoU}^{L1} \times \mathbbm{1}_{obj} + \lambda_{obj}^{CE} \mathcal{L}_{obj}^{CE}, \\
    \mathcal{L}_{Box} &= \mathcal{L}_{Mask} + \mathcal{L}_{box} \\
    &= \mathcal{L}_{Mask} + \left [ \lambda_{box}^{ciou} \mathcal{L}_{box}^{ciou} + \lambda_{box}^{L1} \mathcal{L}_{box}^{L1} \right ] \times \mathbbm{1}_{obj} , \\
    \mathcal{L}_{Point} &= \mathcal{L}_{Mask} + \mathcal{L}_{point} \\
    &= \mathcal{L}_{Mask} + \lambda_{point}^{L1} \mathcal{L}_{point}^{L1}({GT}_{point}, {O}_{point}) \times \mathbbm{1}_{obj},  \\
\end{aligned}
\label{eq:loss}
\end{equation}
where $\mathbbm{1}_{obj}$ denotes we supervise task and IoU prediction only if the object exists, and $\lambda$ represents the weights of different losses.
Table~\ref{tab:app:train} shows the detailed training settings.

\textbf{Inference.}
We conduct all benchmarking experiments on a single A100 GPU using PyTorch 2.5.1 and CUDA 12.1, under automatic mixed precision with bfloat16.
We inference all three tasks following the \emph{fully online inference setting}, i.e., all operations in the current frame can not see the future and only the ground-truth in the first frame is given as a prompt for each target object at the beginning of the sequence without any correction input in the subsequent frames.
For mask tracking task (VOS), we first give each object the ground-truth mask in the first frame and make mask predictions for each object independently and in parallel.
In the multi-object scenario, we merge the per-object logits into a single mask by simply fusing the mask logits based on their values.
For the box tracking task (SOT), the bounding box prediction of the object can be obtained directly from the corner head.
In case of the point tracking task, we replace the prompt with the ground-truth point coordinates and an additional generated mask in Gaussian form, and use the argmax operation to obtain the point coordinates from the mask prediction.
Note that our model is a neat tracker where inference is performed on the complete current frame \textit{without any post-processing strategies}. For example, the centre crop operation, a widely used operation in the SOT task, is able to pre-crop the current frame according to the location in the previous frame, avoiding some incorrect tracking.

\begin{figure*}[t]
    \centering
    \includegraphics[width=\linewidth]{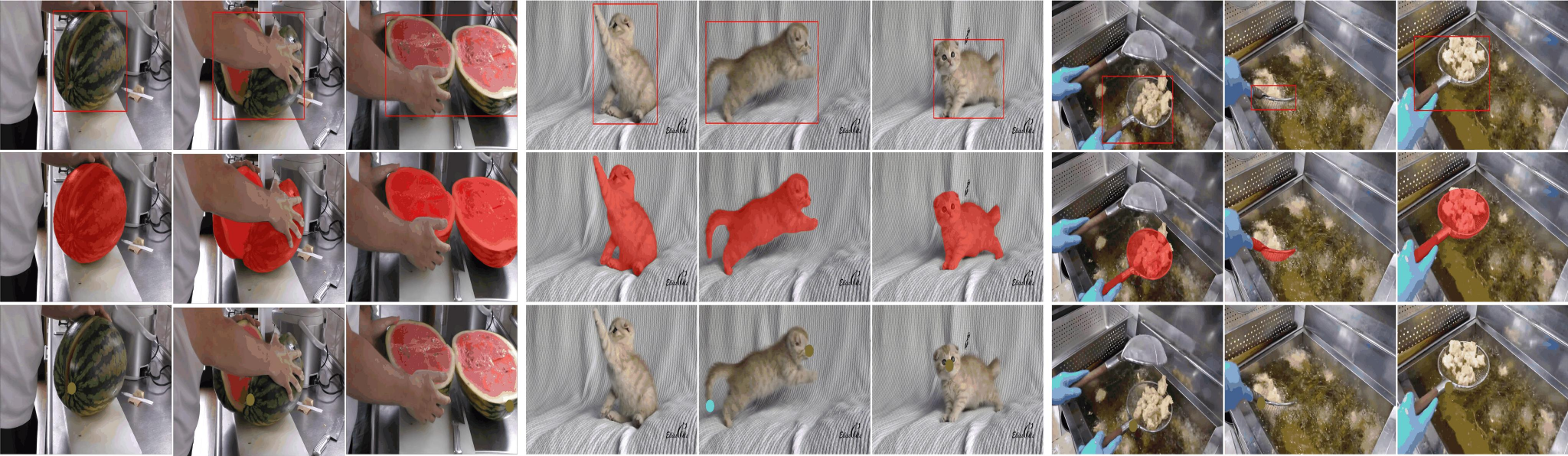}
    \par\vspace{1mm}
    \includegraphics[width=\linewidth]{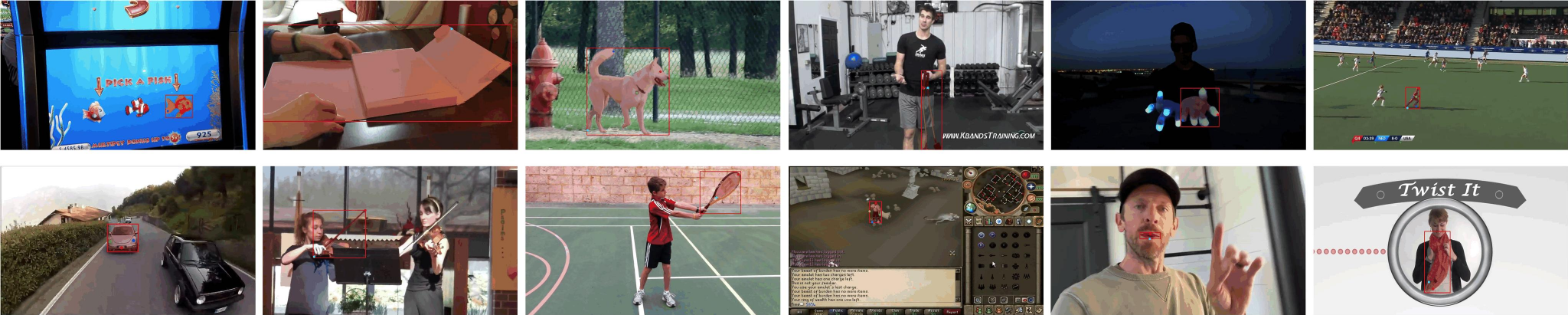}
    \vspace{-3mm}
    \caption{Examples of the Tracking-Any-Granularity dataset. Each sequence is annotated at three granularities: bounding boxes for object-level localization, segmentation masks for pixel-level shape, and key points for fine-grained target tracking.}
    \label{fig:data-example}
    \vspace{-4mm}
\end{figure*}

\begin{table}[t]
    \caption{The interval and number of \textbf{manual annotations} in various phases.}
    \vspace{-1mm}
    \label{tab:data_ta_static_1}
    \centering
    \setlength{\tabcolsep}{1pt}
    \renewcommand{\arraystretch}{1}
    {
    \fontsize{7pt}{10pt}\selectfont
    \begin{tabular}{c|cccccc}
\toprule
TAG &  Videos &   Interval &   Points &   Boxes/Masks &  Total Frames &  Total Len (s)  \\
\midrule
 Phase \textcircled{1} & 1,000 & 1 & 523,137 & 348,715 & 354,625  & 12,540.9  \\
 Phase \textcircled{2} & 2,000 & 10 & 87,708 & 75,923 & 787,643  & 28,164.5   \\
 Phase \textcircled{3} & 3,000 & 20 & 60,809 & 53,917 & 1,058,623  & 39,617.1   \\ 
\midrule
 Total & 6,000 & - & 671,654 & 478,555 & 2,200,891 & 80,322.4   \\ 
\bottomrule
\end{tabular}
    }
    \vspace{-4mm}
\end{table}

\section{Tracking-Any-Granularity Dataset}

We developed a comprehensive dataset for training our unified model, termed \textbf{T}racking-\textbf{A}ny-\textbf{G}ranularity (TAG), with annotations across three granularities: \textit{segmentation masks, bounding boxes, and key points}. Our dataset contains 6,000 high-resolution videos, which preserve fine visual details and enable accurate tracking analysis.
It further encompasses diverse scenes, video sources, and tracked object categories, providing a representative sample of real-world scenarios.
Moreover, TAG includes complex and challenging cases, such as occlusion, motion blur, and other difficult visual conditions, to test the robustness and generalization ability of tracking algorithms.
With comprehensive multi-granularity annotations, a three-phase data engine with model-in-the-loop annotation workflows, and strict multi-stage quality checks, we ensure large-scale, high-quality, and consistent annotations.

\subsection{Target Selection}
Before annotation, we define strict selection criteria for videos, target objects, and target points to ensure data quality and tracking difficulty.
\emph{For videos}, we require continuous clips without camera cuts or scene transitions, clear visuals with identifiable target boundaries, a duration between 10 and 40 seconds excluding static images, and at least one eligible target object.
\emph{For target objects}, we focus on humans, animals, and their parts whose boundaries are distinguishable, which appear clearly in the first frame, contain at least one visible and locatable key point for most of the video while allowing brief occlusions or exits, and remain in motion either actively or passively.
To emphasize challenging tracking scenarios, each target is further required to satisfy at least one difficulty criterion, including rapid motion, high similarity to other objects, occlusion or temporary disappearance, deformation or significant scale/viewpoint changes, and small object size.
\emph{For target points}, we select center points, corner points, or semantically meaningful keypoints such as eyes, hands, or heads; the keypoint must appear in the first frame and is labeled as occluded when it becomes invisible, while points on spherical objects are placed near the center.
We choose keypoints rather than arbitrary points based on two considerations.
From the practical application perspective, downstream tasks such as 3D reconstruction and SLAM usually rely on semantically meaningful keypoints, which offer stronger distinguishing and descriptive capability, and a small number of high-quality keypoints is often sufficient.
From the annotation cost perspective, arbitrary point annotation is prohibitively expensive in real-world videos.
Existing arbitrary-point datasets often rely on indirect annotation sources, such as optical-flow-based trajectory interpolation in RoboTAP or rendering-based point annotations in Kubric and RGB-Stacking.
However, these strategies are not directly applicable to our real-world videos from indoor, outdoor, and wild environments, making it difficult to obtain reliable arbitrary-point annotations automatically.
Therefore, we ask annotators to select target keypoints and manually label them frame by frame, balancing annotation utility and feasibility.

\subsection{Annotation Pipeline}
We designed a coarse-to-fine annotation pipeline to obtain high-quality multi-granularity annotations efficiently and consistently.
The pipeline progressively converts selected videos into aligned point, box, and mask annotations through four steps.
The overall pipeline is demonstrated in Fig.~\ref{fig:anno-pipeline}.

\textbf{1) Video Selection.}
We downloaded a large number of videos from YouTube and instructed annotators to select videos and target objects meeting the above requirements.
This step filters unsuitable clips early and keeps clear, trackable, and challenging targets for annotation.

\textbf{2) Coarse Annotation.}
Annotators first mark key points and tight bounding boxes on target objects.
These annotations provide fine-grained locations and object-level extents, and serve as prompts for generating initial masks.

\textbf{3) Fine Annotation.}
To reduce annotator workload, we use SAM~\cite{kirillov2023seganysam1} to generate rough masks from coarse annotations (points and boxes), so annotators can focus on boundary correction rather than drawing masks from scratch.
Then, annotators refine these masks frame by frame with the following requirements:

\begin{figure*}[t]
    \centering
    \includegraphics[width=\linewidth]{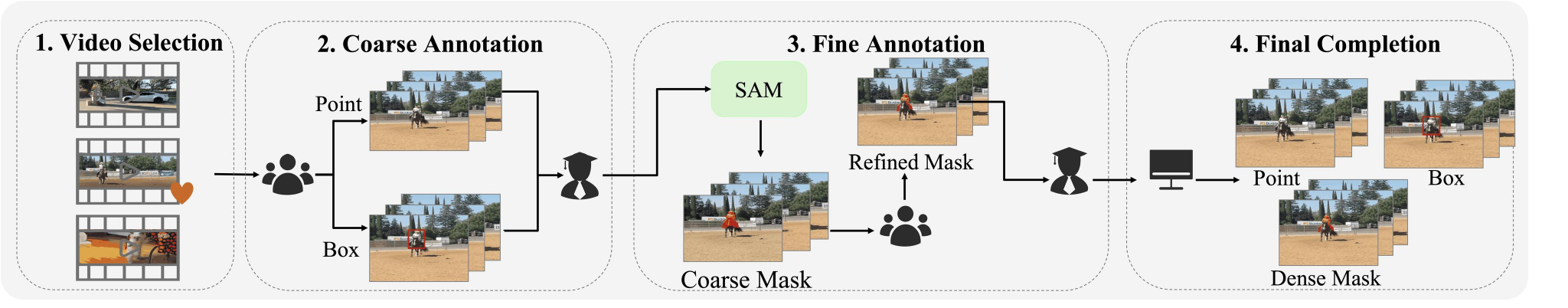}
    \vspace{-1mm}
    \caption{Annotation pipeline of our Tracking-Any-Granularity dataset. The pipeline first selects qualified videos and targets, then performs coarse point and box annotation, uses SAM to produce coarse masks for annotator refinement, and finally conducts expert completion checks to ensure consistency across point, box, and dense mask annotations.}
    \vspace{-1mm}
    \label{fig:anno-pipeline}
\end{figure*}

\begin{table}[t]
\caption{Comparison of Automatic Annotation in Different Annotation Methods.}
\label{app:table:data_engine_ablation}
\centering
\noindent
    \begin{minipage}[t]{\columnwidth}
        \centering
        \textbf{(a) Mask automatic annotation}
        \par
        \vspace{3mm}
        \resizebox{\linewidth}{!}{
            \begin{tabular}{c|cc|ccc|ccc}
\toprule
Type & + 0$_{th}$ & Back. & $\mathcal{J}\&\mathcal{F}^{val}$ & $\mathcal{J}^{val}$ & $\mathcal{F}^{val}$ & $\mathcal{J}\&\mathcal{F}^{test}$ & $\mathcal{J}^{test}$ & $\mathcal{F}^{test}$ \\
\midrule
\#1  & &   & 94.4 & 91.7 & 97.1 & 94.6  & 91.9  & 97.3  \\
\#2  & & \checkmark & 95.0   & 92.4 & 97.6 & 95.0    & 92.4  & 97.6  \\
\#3  & \checkmark  &   & 94.5 & 91.8 & 97.2 & 94.7  & 92.0 & 97.4  \\
\rowcolor[rgb]{0.929,0.902,0.973}
\textbf{\#4}  & \checkmark  & \checkmark & 95.0 & 92.4 & 97.6 & 95.0    & 92.4  & 97.7 \\
\bottomrule
\end{tabular}

        }
    \end{minipage}
    \vspace{2mm}

    \begin{minipage}[t]{\columnwidth}
        \centering
        \textbf{(b) Box automatic annotation}
        \par
        \vspace{3mm}
        \resizebox{\linewidth}{!}{
            \begin{tabular}{c|cc|ccc|ccc}
\toprule
Type & + 0$_{th}$ & Back. & AUC$^{val}$  & $P_{Norm}^{val}$   & $P^{val}$ & AUC$^{test}$  & $P_{Norm}^{test}$   & $P^{test}$ \\
\midrule
\rowcolor[rgb]{0.929,0.902,0.973}
\textbf{\#1} &  & & 84.1    & 94.1  & 91.6  & 84.8 & 94.8   & 91.2   \\
\#2 &  & \checkmark & 83.7    & 93.9  & 91.3  & 84.5 & 94.7   & 91.0   \\
\#3 & \checkmark  & & 83.9    & 93.8  & 91.3  & 84.7 & 94.5   & 91.0   \\
\#4 & \checkmark  & \checkmark & 83.6    & 93.6  & 91.1  & 84.4 & 94.4   & 90.9  \\
\bottomrule
\end{tabular}

        }
    \end{minipage}
    \vspace{2mm}

    \begin{minipage}[t]{\columnwidth}
        \centering
        \textbf{(c) Point automatic annotation}
        \par
        \vspace{3mm}
        \resizebox{0.88\linewidth}{!}{
            \begin{tabular}{c|ccc|cc|cc}
\toprule
Type & + 0$_{th}$ & Back. & Vis. & OA$^{val}$ & AJ$^{val}$ & OA$^{test}$ & AJ$^{test}$ \\
\midrule
\#1 &  &   & -   & 89.0  & 61.7  & 88.7 & 61.3 \\
\rowcolor[rgb]{0.929,0.902,0.973}
\textbf{\#2} & \checkmark   &   & -   & 89.0  & 62.3   & 88.8 & 61.6 \\
\#3 &  & \checkmark  & OR  & 89.3  & 58.1   & 89.6 & 57.3   \\
\#4 &  & \checkmark  & AND & 88.2 & 62.1  & 87.7 & 61.1 \\
\#5 & \checkmark   & \checkmark  & OR  & 89.4  & 58.8   & 89.6 & 57.7   \\
\#6 & \checkmark   & \checkmark  & AND & 88.4 & 62.6  & 87.8 & 61.4  \\
\bottomrule
\end{tabular}
        }
    \end{minipage}
    \vspace{2mm}

    \begin{minipage}[t]{\columnwidth}
        \centering
        \textbf{(d) Mask \& Box Visible automatic annotation}
        \par
        \vspace{3mm}
        {
            \setlength{\tabcolsep}{1.2mm}
            \resizebox{\linewidth}{!}{
                \begin{tabular}{c|ccc|cccc|cccc}
\toprule
Type & + 0$_{th}$ & Backward & Visible & Acc.$^{val}$ & Precision$^{val}$ & Recall$^{val}$ & F1$^{val}$ & Acc.$^{test}$ & Precision$^{test}$ & Recall$^{test}$ & F1$^{test}$ \\
\midrule
\#01 &  &  & AND & 98.47 & 99.52  & 98.86 & 99.12 & 98.89 & 99.54  & 99.27 & 99.38 \\
\#02 &  &  & OR & 98.84 & 98.85  & 99.90 & 99.35 & 99.01 & 99.06  & 99.89 & 99.45 \\
\#03 &  &  & SOT & 98.19 & 98.96  & 99.13 & 98.96 & 98.68 & 99.10  & 99.50 & 99.26 \\
\#04 & \multirow{-4}{*}{} & \multirow{-4}{*}{}   & VOS & 99.12 & 99.42  & 99.63 & 99.51 & 99.22 & 99.49  & 99.66 & 99.56 \\
\midrule
\#05 &  &  & AND & 98.72 & 99.65  & 98.98 & 99.27 & 98.94 & 99.70  & 99.15 & 99.40 \\
\#06 &  &  & OR & 98.71 & 98.65  & 99.98 & 99.27 & 98.91 & 98.96  & 99.90 & 99.40 \\
\#07 &  &  & SOT & 98.36 & 98.70  & 99.58 & 99.07 & 98.80 & 98.96  & 99.78 & 99.34 \\
\#08 & \multirow{-4}{*}{} & \multirow{-4}{*}{\checkmark}   & VOS & 99.08 & 99.60  & 99.38 & 99.48 & 99.05 & 99.70  & 99.27 & 99.46 \\
\midrule
\#09 &  &  & AND & 98.49 & 99.52  & 98.89 & 99.13 & 98.96 & 99.56  & 99.32 & 99.42 \\
\#10 &  &  & OR & 98.84 & 98.83  & 99.92 & 99.34 & 99.01 & 99.04  & 99.91 & 99.45 \\
\#11 &  &  & SOT & 98.21 & 98.94  & 99.17 & 98.97 & 98.68 & 99.08  & 99.52 & 99.26 \\
\#12 & \multirow{-4}{*}{\checkmark} & \multirow{-4}{*}{}   & VOS & 99.12 & 99.41  & 99.63 & 99.51 & 99.30 & 99.51  & 99.72 & 99.61 \\
\midrule
\#13 &  &  & AND & 98.79 & 99.69  & 99.02 & 99.31 & 99.03 & 99.70  & 99.23 & 99.45 \\
\#14 &  &  & OR & 98.73 & 98.67  & 99.98 & 99.28 & 98.91 & 98.94  & 99.92 & 99.40 \\
\#15 &  &  & SOT & 98.37 & 98.72  & 99.57 & 99.08 & 98.78 & 98.94  & 99.78 & 99.32 \\
\rowcolor[rgb]{0.929,0.902,0.973}
\textbf{\#16} & \multirow{-4}{*}{\checkmark} & \multirow{-4}{*}{\checkmark}   & VOS & 99.14 & 99.64  & 99.42 & 99.52 & 99.15 & 99.69  & 99.37 & 99.52 \\
\bottomrule
\end{tabular}

            }
        }
    \end{minipage}
\end{table}

\noindent\begingroup
\setlength{\fboxsep}{4pt}
\fbox{%
\begin{minipage}{\dimexpr\columnwidth-2\fboxsep-2\fboxrule\relax}
\begin{itemize}
    \setlength{\itemsep}{0pt}
    \setlength{\parsep}{0pt}
    \setlength{\topsep}{0pt}
    \item Only annotate the visible parts of the present object in each frame.
    \item In cases of motion blur, infer the approximate position based on the previous frame to maintain temporal consistency. Masks in adjacent frames should not differ drastically.
    \item Ignore transparent or semi-transparent watermarks and subtitles when creating masks; masks can directly cover these elements.
    \item Exclude opaque overlays from the mask, such as logos or captions.
    \item For containers holding other objects, do not include the contained objects in the mask.
    \item The mask should tightly fit the object, neither exceeding nor falling short of its boundaries.
    \item Ensure that mask edges are smooth and avoid excessive roughness.
    \item Fill in small internal holes, but preserve natural gaps (such as hollowed-out structures) or occlusions caused by other objects.
    \item If the initial SAM-generated mask is of very poor quality, annotators may clear it entirely and use color tolerance-based selection to manually annotate the object from scratch.
\end{itemize}
\end{minipage}%
}
\endgroup

\textbf{4) Final Completion.}
Experts perform a final review to thoroughly assess the accuracy and consistency of all three types of annotations, ensuring that the labeling meets the required standards and that any discrepancies are identified and corrected, particularly for challenging scenarios such as occlusions and motion blur.

\begin{table*}[t]
    \setlength{\abovecaptionskip}{3pt}
    \setlength{\belowcaptionskip}{3pt}
    \caption{Comparison of our datasets with public datasets of three tracking datasets in terms of videos, duration, and annotations.}
    \setlength{\tabcolsep}{1mm}
    \label{tab:data_compare_static}
    \centering
    \resizebox{1.0\textwidth}{!}{
    \begin{tabular}{c|cccccccccc}
\toprule
Dataset      & Videos & Total Len. (Avg) & Frames    (Avg) & Resolution  & FPS & Masks    (Avg.) & Boxes    & Points    (Avg.) & Anno. Method   & Motivation \\
\midrule
\multicolumn{11}{c}{\textbf{\emph{Video Object Segmentation}}} \\
DAVIS-2017~\cite{davis17}   & 90     & 5.17 (0.06)      & 6298 (70) & 720p$\sim$4k     & 24  & 13543 (150)     & ×  & ×    & Manual   & Precise labels   \\
BURST~\cite{burst}  & 2914   & 1734 (0.60)      & 624240 (214)    & $\geq$480p & 6   & 600157 (206)    & ×  & ×    & Semi-Automatic & Multi-Task \\
LVOS~\cite{LVOS_V1}   & 220    & 351 (1.60) & 126280 (574)    & 720p  & 6   & 156432 (711)    & ×  & ×    & Manual   & Long-term  \\
LVOS v2~\cite{LVOS_V2}      & 720    & 823 (1.14) & 296401 (412)    & 720p  & 6   & 407945 (567)    & ×  & ×    & Manual   & Large-scale, long-term \\
MOSE~\cite{MOSE}   & 2149   & 443.62 (0.21)    & $\sim$159600 (73)    & 1080p & 6   & 431725 (201)    & ×  & ×    & Semi-Automatic & Complext scenarios     \\
YoutubeVOS-19~\cite{youtube-vos}      & 4453   & 334.8 (0.08)     & 120532 (27)     & 720p  & 6   & 197272 (44)     & ×  & ×    & Manual   & Large-scale      \\
VOST~\cite{vost}   & 713    & 252 (0.35) & 75547 (106)     & 1080p & 5   & 175913 (247)    & ×  & ×    & Semi-Automatic & Object transmission    \\
\midrule
\multicolumn{11}{c}{\textbf{\emph{Single Object Tracking}}} \\
LaSOT~\cite{lasot}  & 1400   & 1950 (1.39)      & 3.52M (2506)    & 720p  & 30  & ×   & 3.52M    & ×    & Manual   & Large-scale, long-term \\
GOT-10k~\cite{got10k}      & 10000  & 2500 (0.25)      & 1.5M (150)      & 720p$\sim$1440p  & 10  & ×   & 1.5M     & ×    & Manual   & Large-scale      \\
TrackingNet~\cite{TrackingNet}  & 30643  & 8400 (0.27)      & 14431266 (471)  & 360p  & 30  & ×   & 14431266 & ×    & Semi-Automatic & Large-scale      \\
UAV123~\cite{uav123} & 123    & 62.5 (0.51)      & 112578 (915)    & 720p  & 30  & ×   & 112578   & ×    & Semi-Automatic & Unmanned aerial vehicles  \\
NfS~\cite{NFS}    & 100    & 26.58 (0.27)     & 383K (3830)     & 720p  & 240 & ×   & 383K     & ×    & Manual   & High Frame Rate  \\
OTB-100~\cite{otb}      & 100    & 32.8 (0.33)      & 59040 (590)     & $\geq$360p & 30  & ×   & 59040    & ×    & Manual   & Real world \\
TNL2K~\cite{wang2021tnl2k}  & 2000   & 691.3 (0.35)     & 1244340 (622)   & 720p  & 30  & ×   & 1244340  & ×    & Manual   & Language-based   \\
VastTrack~\cite{vasttrack}    & 50610  & 11664 (0.23)     & 4.2M (83) & 480p-720p   & 6   & ×   & 4.2M     & ×    & Manual   & Abundant categories    \\
\midrule
\multicolumn{11}{c}{\textbf{\emph{Point Tracking}}} \\
Perception Test~\cite{patraucean2023perception}    & 145    & 55.58 (0.38)     & 100050 (690)    & 720p$\sim$1080p  & 30  & ×   & ×  & 2992705 (20639)  & Manual   & Multi-modal      \\
PointOdyssey~\cite{zheng2023pointodyssey} & 104    & 120 (1.15) & $\sim$216K (2035)    & 540p  & 30  & ×   & ×  & 49B (0.471B)     & Automatic      & Real world, long-term  \\
TAP-Vid Kinetics~\cite{tapvid}   & 1189   & 198.17 (0.17)    & 297250 (250)    & $\geq$720p & 25  & ×   & ×  & 4725959 (3974)   & Semi-Automatic & Abitrary point   \\
TAP-Vid DAVIS~\cite{tapvid}      & 30     & - (-)      & 1999 (66.6)     & 1080p & -   & ×   & ×  & 28824 (960.8)    & Semi-Automatic & Abitrary point   \\
TAP-Vid RGB-Stacking~\cite{tapvid}     & 50     & - (-)      & 12500 (250)     & 256x256     & -   & ×   & ×  & 303436 (6068.7)  & Semi-Automatic & Abitrary point   \\
\midrule
\rowcolor[rgb]{0.929,0.902,0.973}
\textbf{Tracking-Any-Granularity} & 6000   & 1338.7 (0.22)    & 2200891 (367)   & mostly 720p & 30  & 2148716 (358)   & 2148716  & 2640987 (440)    & Semi-Automatic & Any Granularity \\
\bottomrule
\end{tabular}

    }
    \vspace{-2mm}
\end{table*}

\begin{table}[t]
    \setlength{\tabcolsep}{0.5mm}
    \caption{Statistical analysis of video data from our dataset.}
    \vspace{-2mm}
    \label{app:table:data_ta_static_2}
    \centering
    \resizebox{\linewidth}{!}{
        \begin{tabular}{l|cccc}
\toprule
 {} & Average & Medium & Minimum & Maximum                  \\
\midrule
Frame per Video & 366.8 & 295 & 80 & 3,317 \\
Length(s) per Video & 13.39 & 10.9 & 5.3 & 110.9 \\
Video FPS & - & 24 & 10 & 60 \\
\bottomrule
\end{tabular}
    }
    \vspace{-4mm}
\end{table}

\subsection{Data Engine}
As shown in Table~\ref{tab:data_ta_static_1}, the Tracking-Any-Granularity dataset is annotated across three phases:
1) Phase \textcircled{1}: manual annotation of every frame, totaling 1,000 videos;
2) Phase \textcircled{2}: manual annotation of every 10 frames, totaling 2,000 videos;
3) Phase \textcircled{3}: manual annotation of every 20 frames, totaling 3,000 videos.
To increase the size of the dataset while reducing the workload, we adopted a selective annotation strategy in the second and third phases. Instead of manually labeling every video frame, annotators labeled only a subset of frames at varying intervals.
After training the model on both public datasets and the fully labeled data from earlier phases, we leveraged the model to automatically annotate the remaining frames. Specifically, each video was divided into multiple clips, with annotators manually labeling the first and last frames of each clip. The annotation of the first frame in each clip served as the initial target state, enabling the model to infer the target state in the intermediate frames.

To further enhance annotation quality, we introduced two optional refinement methods: (1) performing backward tracking and fusing the results with those from forward tracking, and (2) since the target may be absent in some annotated frames, using the first frame of the entire video, which is guaranteed to contain the target, as an additional reference state alongside the first frame of each clip.
We evaluated these enhancement methods on the Phase 1 validation and test set to determine the inference setting for each tracking task, as shown in Table~\ref{app:table:data_engine_ablation}.
Specifically, the four subtables represent the evaluation outcomes for the VOS, SOT, PT, and object existence prediction tasks under various settings, respectively.
By comparing the results across different settings, we select the configuration highlighted in the gray row as the inference setting for each task.

Although the data engine's concept is inspired by SAM 2, there are significant differences. The data engine of SAM 2 requires repeated manual correction and re-tracking, which hinders progress. Instead, we decouple manual and automated processes using interval annotations, enhancing efficiency while avoiding error accumulation.

\begin{figure*}[t]
    \centering
    \includegraphics[width=\textwidth]{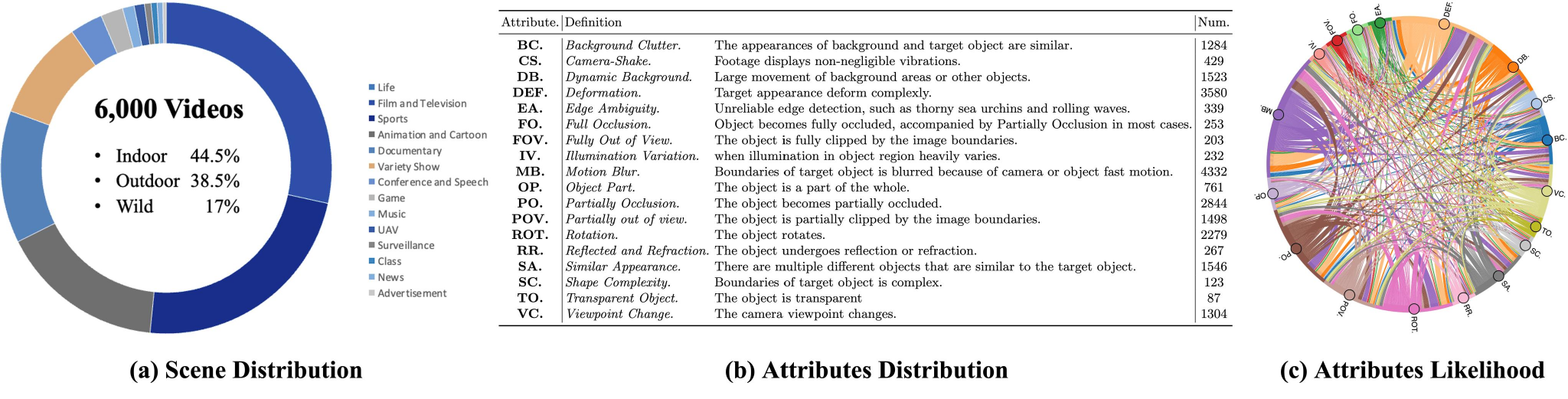}
    \vspace{-8mm}
    \caption{Attribute analysis of video data from our dataset. The figure reports the number of sequences associated with each challenging attribute, covering 18 categories such as motion blur, deformation, occlusion, scale variation, and viewpoint change. Since attributes are not mutually exclusive, one sequence may be assigned to multiple categories.}
    \label{fig:attr}
    \vspace{-4mm}
\end{figure*}

\subsection{Tracking-Any-Granularity Dataset}

Compared with existing datasets, our challenging real-world Tracking-Any-Granularity dataset stands out as the only one providing annotations at all three granularities simultaneously.
It comprises 6,000 videos, each annotated with masks, boxes, and points.
The majority of the videos have a resolution of $1280\times 720$, with 398 exceptions; their duration ranges from $5.3$ seconds to $110.9$ seconds, and the frame count ranges from $80$ to $3,317$ frames.
In total, TAG contains $2.2$ million frames and $1,338.7$ minutes of videos.
More detailed statistics are shown in Table~\ref{app:table:data_ta_static_2}, and annotation examples are shown in Fig.~\ref{fig:data-example}.
We compare our dataset with numerous public datasets in Table~\ref{tab:data_compare_static}, showing our dataset contains significantly more videos and annotations than they do, creating a substantial resource for multi-granularity tracking research.

\textbf{Scene and Attribute.}
To enable a more comprehensive analysis of tracking approaches, it is critically important to identify video scenes and attributes of our dataset.
Furthermore, we label each sequence with 18 attributes that represent various video challenges, and the detailed definitions and counts are shown in Fig.~\ref{fig:attr}.
It is worth noting that these attributes are not mutually exclusive, and a single video may contain multiple challenges.
Motion Blur, Deformation, and Partial Occlusion are the most common challenges in our dataset, demonstrating its high level of difficulty.

\begin{table*}[t]
    \caption{State-of-the-art comparison on Video Object Segmentation Task.}
    \vspace{-2mm}
    \label{tab:sota_vos}
    \centering
    \resizebox{1.0\textwidth}{!}{
    \begin{tabular}{l|ccc|ccc|ccc|ccc|ccc|cc|ccccc}
\toprule
  & \multicolumn{3}{c|}{ BURST$^{test}$}  & \multicolumn{3}{c|}{ LVOS$_{v2}^{val}$}  & \multicolumn{3}{c|}{ MOSE$^{val}$} & \multicolumn{3}{c|}{ TAG$_{VOS}^{test}$} & \multicolumn{3}{c|}{ TAG$_{VOS}^{val}$} & \multicolumn{2}{c|}{ VOST$_{10fps}^{val}$} & \multicolumn{5}{c}{ YoutubeVOS$_{2019}^{val}$}    \\
\multirow{-2}{*}{Methods}  &  H$_{all}$ &  H$_{com}$ &  H$_{unc}$ & $\mathcal{J}\&\mathcal{F}$ & $\mathcal{J}$ & $\mathcal{F}$   & $\mathcal{J}\&\mathcal{F}$  & $\mathcal{J}$  & $\mathcal{F}$ & $\mathcal{J}\&\mathcal{F}$    & $\mathcal{J}$    & $\mathcal{F}$     & $\mathcal{J}\&\mathcal{F}$    & $\mathcal{J}$    & $\mathcal{F}$   & $\mathcal{J}$  & $\mathcal{J}_{tr}$ & Overall & $\mathcal{J}_{seen}$ & $\mathcal{F}_{seen}$ & $\mathcal{J}_{unseen}$ & $\mathcal{F}_{unseen}$ \\
\midrule
STCN~\cite{stcn} & 45.8   & 45.6  & 45.9 & 62.0  & 58.5  & 65.4   & 50.1    & 46.1 & 54.1   & 70.4  & 65.9  & 75.0   & 76.2  & 72.2  & 80.2  & 32.7 & 22.7  & 82.8 & 81.3 & 85.6 & 78.3   & 86.0   \\
AOT-SwinB~\cite{aot}   & 54.9   & 55.3  & 54.8 & 73.8  & 70.1  & 77.4    & 60.2    & 56.2 & 64.1   & 78.1  & 73.1  & 83.2   & 80.9  & 76.4  & 85.4  & 39.3 & 29.4  & 85.3 & 84.6 & 89.5 & 79.3   & 87.7   \\
DeAOT-SwinB~\cite{deaot} & \underline{58.5}   & 58.0  & \underline{58.6} & 72.1  & 68.5  & 75.8   & 61.7    & 57.6 & 65.9   & \underline{79.6}  & 74.8  & \underline{84.4}   & 81.6  & 77.3  & 85.9  & 43.0 & 29.4  & 86.4 & 85.4 & 90.3 & 80.6   & 89.3   \\
XMem~\cite{xmem} & 52.3   & 51.9  & 52.3 & 64.7  & 61.8  & 67.5  & 56.3    & 52.1 & 60.5   & 74.4  & 70.1  & 78.6   & 75.7  & 71.8  & 79.6  & 37.9 & 25.0  & 85.5 & 84.3 & 88.6 & 80.5   & 88.7   \\
DEVA~\cite{deva} & 56.1   & 55.4  & 56.2 & 72.2  & 68.6  & 75.9  & 66.4    & 62.2 & 70.6   & 77.9  & 73.1  & 82.6   & 82.1  & 78.0  & 86.1 & 40.9 & 27.3  & 86.3 & 85.3 & 89.8 & 80.6   & 89.2   \\
Cutie-base+~\cite{cutie} & 55.8   & 55.6  & 55.8 & 71.4  & 68.6  & 74.3  & 66.2    & 62.3 & 70.1   & 79.0  & \underline{75.0}  & 83.0   & \underline{83.8}  & \underline{80.0}  & \underline{87.7}  & 44.7 & \textbf{32.7}  & 86.9 & \textbf{86.2} & \textbf{90.7} & 81.6   & 89.2   \\
OneVOS$_{\text{w/MOSE}}$ ~\cite{li2024onevos}  & 57.9   & \underline{59.4}  & 57.6 & \underline{74.7}  & \underline{71.1}  & \underline{78.3}  & 62.2    & 57.9 & 66.6   & 79.3  & 74.3  & 84.3   & 82.4  & 78.0  & 86.7  & \underline{44.9} & 29.0  & 86.3 & 84.9 & 89.9 & 81.1   & 89.4   \\
JointFormer~\cite{jointformer}  & -   & -  & - & 71.7  & 68.8  & 74.7  & \underline{69.7}    & \underline{65.8} & \underline{73.6}   & 76.6 & 72.8 & 80.5 & 79.1 & 75.5 & 82.7 & -  & - & \underline{87.0} & 86.1 & 90.6 & \underline{82.0} & \underline{89.5}   \\
\rowcolor[rgb]{0.929,0.902,0.973}
Ours & \textbf{66.4}   & \textbf{66.5}  & \textbf{66.4} & \textbf{82.2}  & \textbf{78.7}  & \textbf{85.7}  & \textbf{74.6}    & \textbf{70.6} & \textbf{78.6}   & \textbf{87.4}  & \textbf{84.2}  & \textbf{90.7} & \textbf{87.9}  & \textbf{84.9}  & \textbf{90.9} & \textbf{45.2} & \underline{25.6}  & \textbf{87.1} & \underline{85.8} & \underline{82.5} & \textbf{90.0} & \textbf{90.3}  \\
\bottomrule
\end{tabular}

    }
    \vspace{-2mm}
\end{table*}

\begin{table*}[t]
  \caption{State-of-the-art comparison on Single Object Tracking Task.}
    \vspace{-1mm}
  \label{tab:sota_sot}
  \centering
    \resizebox{1.0\textwidth}{!}{
    \begin{tabular}{l|ccc|ccc|ccc|ccc|ccc}
\toprule
\multirow{2}{*}{Methods} & \multicolumn{3}{c|}{GOT10k$^{test}$} & \multicolumn{3}{c|}{TAG$_{SOT}^{test}$} & \multicolumn{3}{c|}{TAG$_{SOT}^{val}$} & \multicolumn{3}{c|}{TrackingNet} & \multicolumn{3}{c}{VastTrack} \\ \cline{2-16}
& AO & $SR_{0.5}$  & $SR_{0.75}$ & AUC  & $P_{Norm}$   & $P$   & AUC & $P_{Norm}$   & $P$   & AUC  & $P_{Norm}$ & $P$ & AUC & $P_{Norm}$   & $P$ \\
\midrule
OSTrack$_{\text{256}}$~\cite{ye2022joint_OStrack}           & 74.8   & 84.2        & 72.2         & 68.3        & 77.0            & 68.1      & 66.7        & 75.8            & 64.7     & 83.1     & 87.8         & 81.9  & 33.8    & 41.3        & 31.1  \\
OSTrack$_{\text{384}}$~\cite{ye2022joint_OStrack}           & 74.8   & 84.4        & 72.7         & 69.7        & 78.8            & 69.9      & 68.3        & 77.1            & 66.2     & 83.8     & 88.5         & 83.2  & 33.7    & 40.8        & 31.4  \\
SimTrack$_{\text{224}}$~\cite{pauwels2015simtrack}        & 71.1   & 80.5        & 68.1         & 64.1        & 72.4            & 60.5      & 65.8        & 73.7            & 63.7     & 82.3     & 86.7         & 80.2  & 34.5    & 40.5        & 30.4  \\
DropTrack$_{\text{384}}$~\cite{wu2023dropmae_droptrack}         & 76.8   & 86.9        & 74.4         & 71.1        & 80.5            & 72.1      & 70.8        & 80.4            & 69.4     & 83.8     & 88.5         & 83.1  & 37.5    & 45.9        & 36.4  \\
GRM$_{\text{256}}$~\cite{Gao_2023_CVPR_GRM}               & 73.1   & 82.3        & 71.4         & 69.1        & 77.4            & 69.1      & 68.5        & 77.3            & 66.5     & 83.8     & 88.5         & 83.1  & 34.6    & 42.2        & 32.3  \\
MixViT$_{\text{288}}^{\text{ConvMAE}}$~\cite{MixFormer}  & 72.1   & 80.9        & 70.5         & 69.7        & 78.2            & 70.2      & 66.3        & 74.9            & 64.4     & 83.7     & 88.6         & 82.8  & 36.4    & 44.0        & 34.9  \\
SeqTrack$_{\text{256}}$~\cite{chen2023seqtrack}         & 76.1   & 85.5        & 74.7         & 68.4        & 78.0            & 69.0      & 67.0        & 76.5            & 64.7     & 83.2     & 88.3         & 82.3  & 34.8    & 43.6        & 33.6  \\
SeqTrack$_{\text{384}}$~\cite{chen2023seqtrack}         & 77.0   & 85.8        & 76.1         & 69.8        & 79.4            & 71.5      & 68.5        & 78.2            & 67.8     & 84.0     & 89.0         & 83.7  & 35.8    & 44.8        & 35.3  \\
ARTrack$_{\text{256}}$~\cite{wei2023autoregressiveARtrack}           & 76.8   & 85.8        & 75.7         & 71.1        & 78.7            & 70.9      & 69.9        & 76.9            & 67.1     & 84.4     & 88.8         & 83.5  & 35.7    & 42.1        & 32.4  \\
ARTrack-V2$_{\text{256}}$~\cite{bai2024artrackv2}        & 76.3   & 85.5        & 74.3         & 71.8        & 79.5            & 71.9      & 70.2        & 78.0            & 68.3     & 84.2     & 88.8         & 83.4  & 37.0    & 44.5        & 34.8  \\
ARTrack-V2$_{\text{384}}$~\cite{bai2024artrackv2}        & -      & -           & -            & -           & -               & -         & -           & -               & -        & 85.7     & 89.8         & 85.5  & -           & -           & -           \\
ROMTrack$_{\text{256}}$~\cite{cai2023robust_romtrack}        & 74.9   & 85.0        & 73.5         & 70.0        & 79.1            & 70.2      & 68.8        & 77.7            & 67.5     & 83.7     & 88.4         & 82.8  & 35.4    & 43.6        & 33.5  \\
ROMTrack$_{\text{384}}$~\cite{cai2023robust_romtrack}        & 75.6   & 85.4        & 73.7         & 71.3        & 80.8            & 72.8      & 69.2        & 77.8            & 68.5     & 83.8     & 88.5         & 83.1  & 37.1    & 45.5        & 36.2  \\
HIPTrack$_{\text{384}}$~\cite{cai2024hiptrack}          & 78.2   & 88.5        & 76.6         & 71.4        & 81.0            & 72.5      & 72.0        & 82.2            & 71.0     & 84.5     & 89.2         & 83.7  & 38.6    & 46.3        & 36.8  \\
LoRAT$_{\text{224}}$~\cite{lin2024tracking_lorat}           & 75.1   & 84.8        & 74.4         & 70.5        & 79.7            & 68.7      & 72.7        & 82.2            & 74.4     & 83.6     & 88.0         & 82.2  & 38.7    & 37.8        & 41.1  \\
LoRAT$_{\text{378}}$~\cite{lin2024tracking_lorat}           & 77.7   & 87.5        & 77.5         & 73.3        & 83.1            & 73.4      & 73.2        & 82.7            & 76.4     & 84.3     & 88.5         & 83.2  & 40.4    & 40.7        & 43.4  \\
SUTrack$_{\text{224}}$~\cite{SUTrack}              & 77.9   & 87.5        & \textbf{78.5}         & \underline{75.5}        & \textbf{85.9}            & \underline{78.7}      & \underline{75.0}        & \underline{85.2}            & \underline{77.1}     & 85.7     & \underline{90.3}         & 85.1  & \underline{44.2}    & \underline{53.8}        & \underline{44.8}  \\
SAMURAI~\cite{yang2024samurai}                  & \underline{79.6}   & \textbf{90.8}        & 72.9         & -           & -               & -         & -           & -               & -        & 85.3     & -            & -     & -       & -           & -     \\
DreamTrack$_{\text{256}}$~\cite{DreamTrack}           & -      & -           & -            & -           & -               & -         & -           & -               & -        & 85.8     & 90.0         & 85.3  & -       & -           & -     \\
DreamTrack$_{\text{384}}$~\cite{DreamTrack}           & -      & -           & -            & -           & -               & -         & -           & -               & -        & \textbf{86.5}     & \textbf{90.6}         & \underline{85.9}  & -       & -           & -     \\
\rowcolor[rgb]{0.929,0.902,0.973}
Ours                     & \textbf{80.7}   & \underline{89.7}        & \underline{77.8}         & \textbf{78.0}        & \underline{85.7}            & \textbf{81.5}      & \textbf{78.2}        & \textbf{86.2}            & \textbf{82.0}     & \underline{86.0}     & 90.1         & \textbf{87.3}  & \textbf{55.0}    & \textbf{65.5}        & \textbf{60.4} \\
\bottomrule
\end{tabular}

    }
    \vspace{-2mm}
\end{table*}

\begin{table*}[t]
  \caption{State-of-the-art comparison on Point Tracking Task. \textcolor{gray}{model} and model$^{\dag}$ means original offline trackers and modified online trackers.}
    \vspace{-1mm}
  \label{tab:sota_pt}
  \centering
    \resizebox{1.0\textwidth}{!}{
    \begin{tabular}{l|cccc|cc||l|cccc|cc}
\toprule
\multirow{2}{*}{Method} & \multicolumn{4}{c|}{Key Point} & \multicolumn{2}{c||}{Arbitrary Point} & \multirow{2}{*}{Method} & \multicolumn{4}{c|}{Key Point} & \multicolumn{2}{c}{Arbitrary Point} \\
& BADJA & Perc.$^{val}$ & TAG$_{PT}^{test}$ & TAG$_{PT}^{val}$ & DAVIS & RGB & & BADJA & Perc.$^{val}$ & TAG$_{PT}^{test}$ & TAG$_{PT}^{val}$ & DAVIS & RGB \\
\midrule
\textcolor{gray}{pips}~\cite{harley2022pips} & \textcolor{gray}{64.2} & \textcolor{gray}{41.5} & \textcolor{gray}{19.0} & \textcolor{gray}{19.8} & \textcolor{gray}{40.9} & \textcolor{gray}{28.5} & \textcolor{gray}{TAPTR}~\cite{TAPTR} & \textcolor{gray}{69.1} & \textcolor{gray}{59.4} & \textcolor{gray}{23.7} & \textcolor{gray}{23.8} & \textcolor{gray}{61.2} & \textcolor{gray}{58.0} \\
pips$^{\dag}$~\cite{harley2022pips} & 56.9 & 27.2 & 12.3 & 12.2 & 26.3 & 20.8 & TAPTR$^{\dag}$~\cite{TAPTR} & 63.0 & 48.9 & 20.4 & 19.0 & 52.3 & 39.4 \\
\textcolor{gray}{pips++}~\cite{zheng2023pointodyssey} & \textcolor{gray}{56.6} & \textcolor{gray}{62.6} & \textcolor{gray}{20.9} & \textcolor{gray}{23.1} & \textcolor{gray}{59.8} & \textcolor{gray}{58.5} & \textcolor{gray}{TAPIR}~\cite{doersch2023tapir} & \textcolor{gray}{64.2} & \textcolor{gray}{59.6} & \textcolor{gray}{21.3} & \textcolor{gray}{24.6} & \textcolor{gray}{56.8} & \textcolor{gray}{50.4} \\
\textcolor{gray}{CoTracker}~\cite{cotrackerv1} & \textcolor{gray}{65.3} & \textcolor{gray}{59.8} & \textcolor{gray}{23.3} & \textcolor{gray}{22.3} & \textcolor{gray}{60.9} & \textcolor{gray}{63.1} & \textcolor{gray}{LocoTrack}~\cite{LocoTrack} & \textcolor{gray}{68.7} & \textcolor{gray}{67.1} & \textcolor{gray}{25.2} & \textcolor{gray}{30.2} & \textcolor{gray}{62.7} & \textcolor{gray}{69.2} \\
CoTracker$^{\dag}$~\cite{cotrackerv1} & 55.1 & 48.9 & 18.8 & 18.1 & 50.8 & 46.1 & Track-On~\cite{Track-On} & \underline{69.7} & \textbf{69.7} & 24.8 & \underline{25.8} & \textbf{64.5} & \textbf{64.5} \\
\textcolor{gray}{CoTracker3}~\cite{cotrackerv3} & \textcolor{gray}{72.7} & \textcolor{gray}{71.3} & \textcolor{gray}{29.6} & \textcolor{gray}{29.1} & \textcolor{gray}{65.6} & \textcolor{gray}{70.6} & \cellcolor[rgb]{0.929,0.902,0.973}Ours & \cellcolor[rgb]{0.929,0.902,0.973}\textbf{72.9} & \cellcolor[rgb]{0.929,0.902,0.973}66.2 & \cellcolor[rgb]{0.929,0.902,0.973}\textbf{35.3} & \cellcolor[rgb]{0.929,0.902,0.973}\textbf{37.7} & \cellcolor[rgb]{0.929,0.902,0.973}56.1 & \cellcolor[rgb]{0.929,0.902,0.973}59.0 \\
CoTracker3$^{\dag}$~\cite{cotrackerv3} & 66.3 & \underline{66.3} & \underline{25.8} & 24.9 & \underline{59.2} & \underline{63.6} & & & & & & & \\
\bottomrule
\end{tabular}

    }
    \vspace{-3mm}
\end{table*}
    
\textbf{Dataset Splits.}
As summarized in Table~\ref{tab:data_ta_static_1}, Phase \textcircled{1} contains 1,000 fully annotated videos with dense manual labels on every frame, making it the most reliable source for evaluation splits.
From this phase, we selected 150 validation videos and 150 test videos with stratified sampling based on both category and source, ensuring a balanced distribution and reducing potential bias for fair evaluation across diverse video types.

\section{Experiments}

\begin{figure*}[t]
    \centering
    \begin{minipage}{\linewidth}
        \centering
        \includegraphics[width=\linewidth]{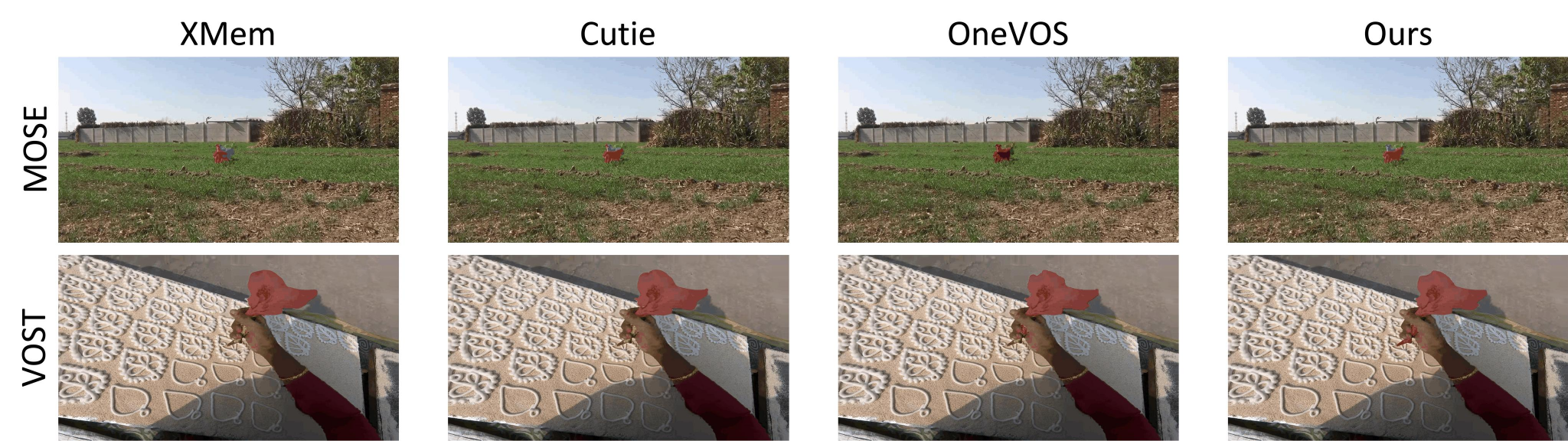}
    \end{minipage}
    \par\vspace{2mm}
    \noindent\makebox[\linewidth]{\leaders\hbox{\rule{4pt}{0.4pt}\hspace{3pt}}\hfill\kern0pt}
    \begin{minipage}{\linewidth}
        \centering
        \includegraphics[width=\linewidth]{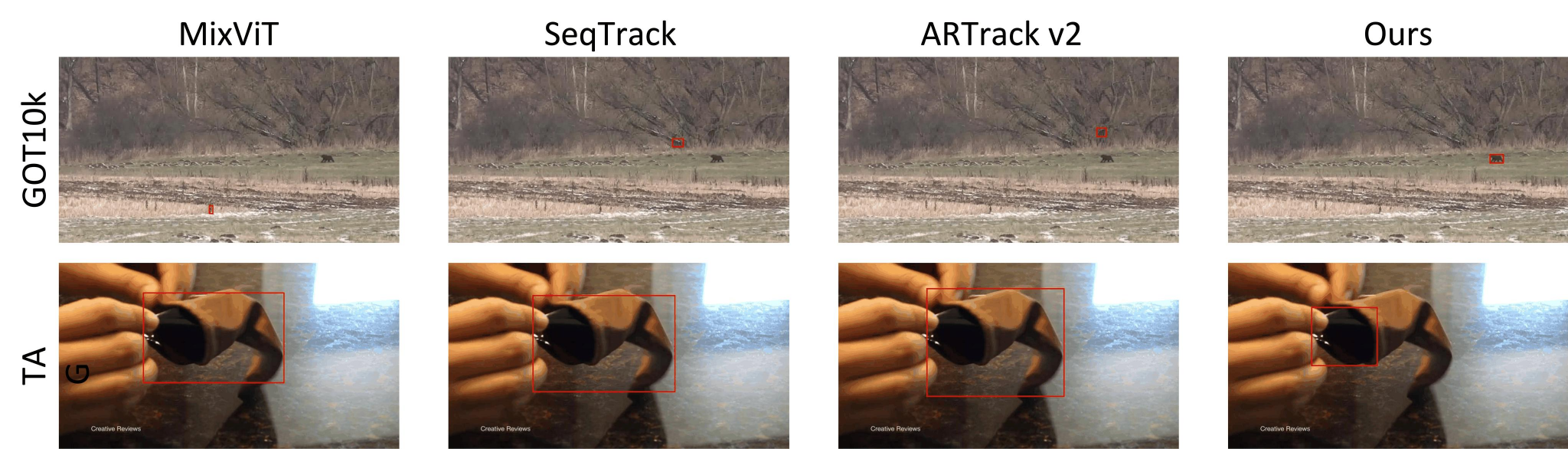}
    \end{minipage}
    \par\vspace{2mm}
    \noindent\makebox[\linewidth]{\leaders\hbox{\rule{4pt}{0.4pt}\hspace{3pt}}\hfill\kern0pt}
    \begin{minipage}{\linewidth}
        \centering
        \includegraphics[width=\linewidth]{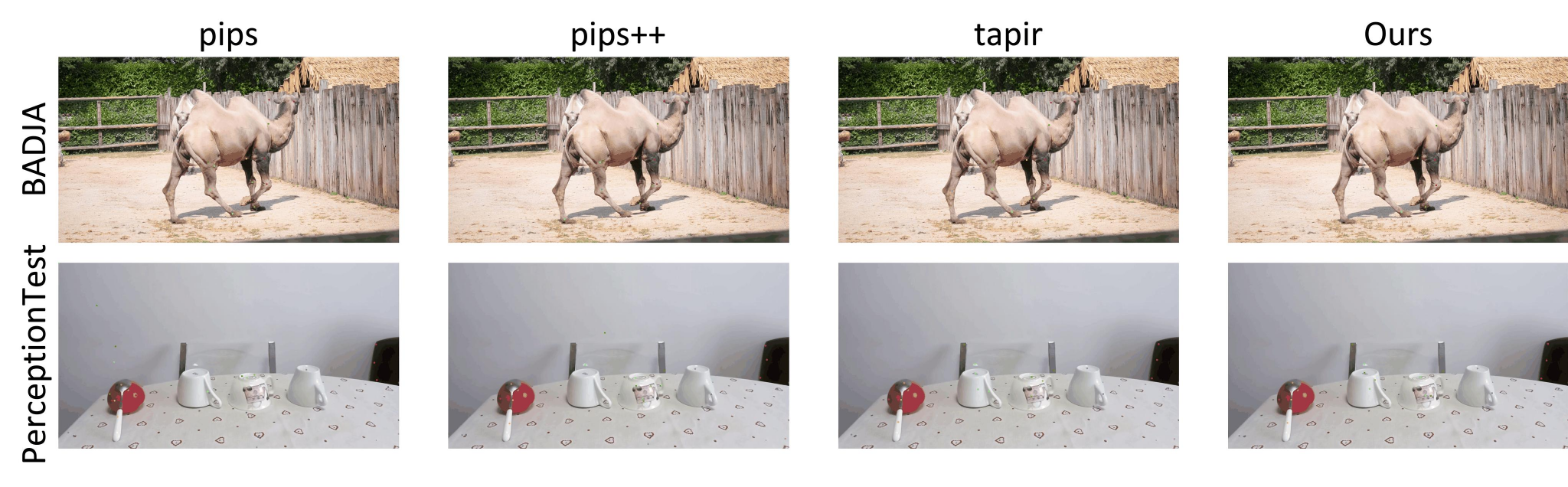}
    \end{minipage}
    \vspace{-6mm}
    \caption{Comparison between our model and various SOTA methods on video tracking benchmarks at three granularities. The three groups show representative results for mask-level video object segmentation, box-level single object tracking, and point-level online point tracking. Across scenarios with deformation, occlusion, fast motion, and appearance changes, SAM 2++ produces more stable target states and accurate localization than task-specific baselines, preserving object boundaries, bounding boxes, and semantic points across frames.}
    \label{fig:app:result_comp}
    \vspace{-4mm}
\end{figure*}

\begin{figure*}[t]
    \centering
    \includegraphics[width=\linewidth]{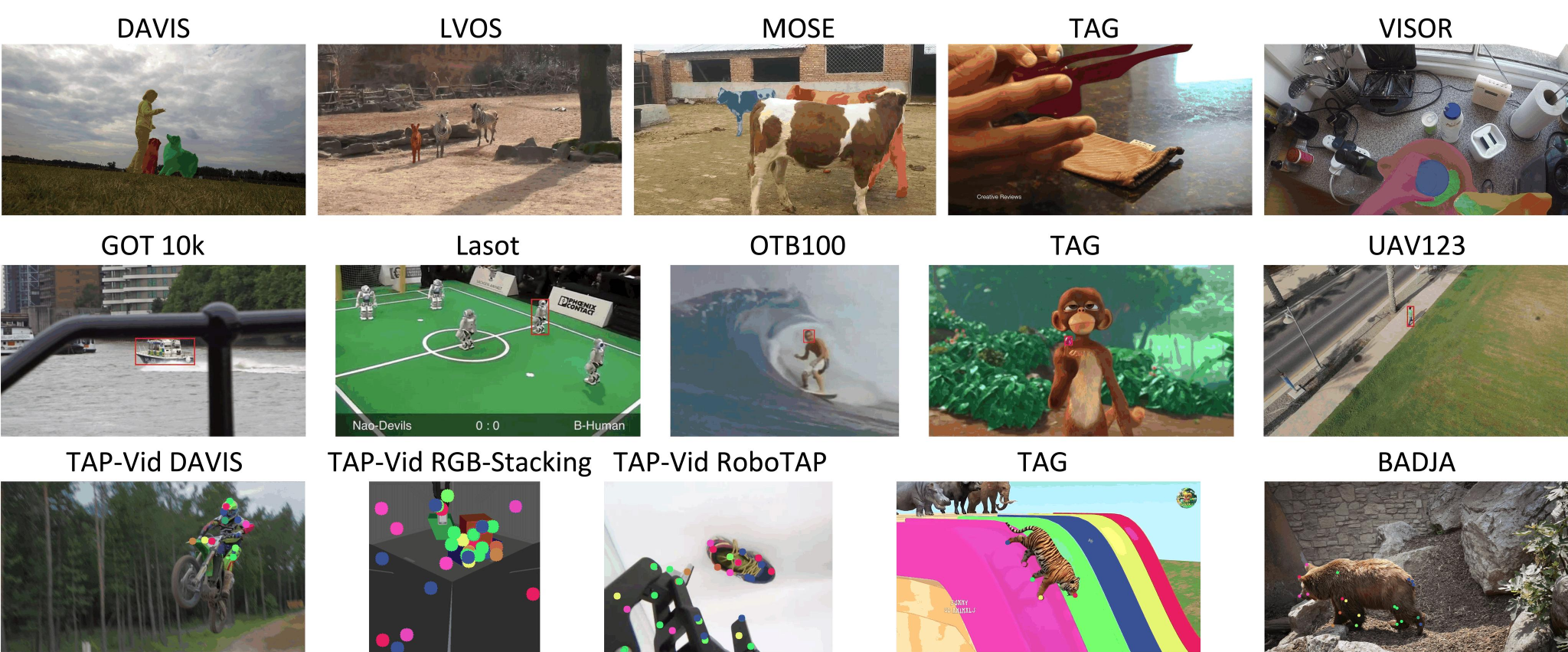}
    \caption{Examples from SAM 2++ results on video benchmarks at various granularities. The same model outputs segmentation masks, bounding boxes, and target points from different prompts, while maintaining consistent predictions across diverse categories and motion patterns.}
    \label{fig:app:model_result}
    \vspace{-3mm}
\end{figure*}

\subsection{Comparison to state-of-the-art on three tasks}

\textbf{Video Object Segmentation.}
The comparisons between our model and previous semi-supervised VOS methods are demonstrated in Table~\ref{tab:sota_vos}, including YoutubeVOS-19~\cite{youtube-vos}, MOSE~\cite{MOSE}, LVOS-v2~\cite{LVOS_V2}, BURST~\cite{burst}, VOST~\cite{vost}, VISOR~\cite{visor}, and our TAG dataset.
For evaluation, we use standard metrics~\cite{vos-benchmark} in most benchmarks: Jaccard index $\mathcal{J}$, contour accuracy $\mathcal{F}$, and their average $\mathcal{J\&F}$.
In the YouTubeVOS benchmark, $\mathcal{J}$ and $\mathcal{F}$ are computed for "seen" and "unseen" categories separately, and $\mathcal{G}$ is the averaged $\mathcal{J\&F}$ for both seen and unseen classes.
In LVOS, the standard deviation $\mathcal{V}$ of the average score of $\mathcal{J}$ and $\mathcal{F}$ assesses temporal stability, while VOST additionally reports $\mathcal{J}_{tr}$ for the last 25\% frames to evaluate robustness after object transformation.
The BURST benchmark is evaluated with Higher Order Tracking Accuracy (HOTA)~\cite{hota}, which balances frame-level detection and temporal association accuracy.
Results show that our model outperforms individual video object segmentation models.

\textbf{Single Object Tracking.}
We compare the performance of our proposed model on three benchmarks in Table~\ref{tab:sota_sot}, including TrackingNet~\cite{TrackingNet}, GOT-10k~\cite{got10k}, TNL2K~\cite{wang2021tnl2k}, VastTrack~\cite{vasttrack} and our TAG dataset, and all compared models are trained on four datasets.
For most benchmarks, we evaluate performance with Area Under the Curve (AUC), normalized precision (${P}_{Norm}$), and precision (${P}$), which measure the average accuracy of center, size, and scale between predicted and ground-truth bounding boxes.
For the GOT-10k benchmark, we choose Average Overlap (AO) and success rate (SR) as indicators, where AO denotes the average overlap between all ground-truth and estimated boxes, and SR measures the percentage of successfully tracked frames whose overlap exceeds a threshold (e.g., 0.5).
Experiments demonstrate that our model consistently outperforms existing methods across nearly all benchmarks.
Crucially, unlike specialized SOT trackers that focus on a single task and employ a pre-cropping process, our framework is an end-to-end, unified tracker.

\textbf{Online Point Tracking.}
We compare our method to prior works in Table~\ref{tab:sota_pt} on four benchmarks, including BADJA~\cite{badja} and Perception Test~\cite{patraucean2023perception} for key point tracking, TAP-Vid~\cite{tapvid} for arbitrary point tracking, and our TAG dataset.
For TAP-Vid~\cite{tapvid}, Perception Test~\cite{patraucean2023perception}, and our TAG dataset, we report Occlusion Accuracy (OA) and Average Jaccard (AJ).
For the BADJA benchmark, we adopt the Percentage of Correct Keypoint-Transfer (PCK-T).
We measure these benchmarks in the `query first' evaluation, which means points appearing in the first frame are used as queries.
However, most of the current methods are \textcolor{gray}{offline trackers}, which process long-temporal window frames or even the entire video to be able to see the future frame, and do not match the online setup required by real applications.
For a fair comparison, we modified their input so that there is no future information inside the window, denoted as model$^{\dag}$, to enable inference online.
Experimental results show a substantial decrease in model performance when the input data is switched from offline to online.
The comparative analysis reveals the effectiveness of our approach on keypoint tracking benchmarks, which surpasses competing models.
Furthermore, although our model is trained on keypoint datasets, it demonstrates generalization capability on arbitrary point tracking datasets.

\textbf{Comparison with TAG model.}
TAG~\cite{harley2024tag} shares our goal of unified mask, box, and point tracking, so we compare the two approaches to clarify our contributions.
First, TAG is an offline model that processes multiple frames as a clip, which differs from the mainstream online frame-by-frame setting, may leak future information, and is mainly applicable to pre-recorded videos. In contrast, our model uses only past information for the current frame, making it suitable for video streams.
Second, TAG converts point coordinates into a \{0, 1\} mask, providing limited target information. Our method combines point coordinates with a (0, 1) Gaussian mask, using precise coordinates for localization and the Gaussian mask to highlight the target in a form consistent with MaskDecoder outputs and MemoryEncoder inputs. For box tracking, TAG converts a box into a square mask, which may confuse the target region with the background.
Third, TAG is trained only on public datasets, while our Data Engine supports both model training and annotation expansion, yielding a large-scale dataset with three granularities and a stronger unified model.
Most importantly, TAG focuses on jointly encoding targets of different granularities in an offline clip-level setting. When moving to the next clip, it still relies on prompts in their original form and must use the last-frame prediction from the previous clip as the next prompt, even when that prediction is unreliable, which can cause error accumulation and makes temporary disappearance difficult to handle. Our online formulation instead requires converting predictions into memory representations for subsequent frames. Memory provides richer target features than raw prompts and, with selective capability and memory diversity, improves tracking stability. To support this across granularities, we introduce \textbf{Task-Adaptive Memory} to unify predictions of different forms into memory representations. Thus, the difference from TAG is not only about task coverage, but also about motivation and core design.
As shown in Table~\ref{app:table:sota_tag}, our model significantly outperforms TAG on all three tasks.

\begin{table}[t]
    \caption{Comparison with TAG Models.}
    \label{app:table:sota_tag}
    \centering
    \begin{tabular}{l|c|cc|c}
\toprule
Methods & VOST & LaTOT & TOTB & CroHD \\
\midrule
TAG~\cite{harley2024tag}  & 31.3 & 35.3 & 74.4 & 57.1 \\
\rowcolor[rgb]{0.929,0.902,0.973}
Ours & \textbf{45.2} & \textbf{37.4} & \textbf{83.8} & \textbf{66.6} \\
\bottomrule
\end{tabular}
    \vspace{-4mm}
\end{table}

\begin{table}[t]
    \caption{Comparison with Unified Models.}
    \label{app:table:sota_unifiedmodel}
    \centering
    \resizebox{1.0\columnwidth}{!}{
        \begin{tabular}{l|ccc|ccc}
\toprule
\multirow{2}{*}{Methods} & \multicolumn{3}{c|}{DAVIS$_{2017}^{val}$} & \multicolumn{3}{c}{TrackingNet}   \\
& $\mathcal{J}\&\mathcal{F}$ & $\mathcal{J}$ & $\mathcal{F}$ & AUC  & $P_{Norm}$ & $P$ \\
\midrule
Unicorn~\cite{Unicorn} & 69.2 & 65.2 & 73.2 & 83.0 & 86.4 & 82.2 \\
UNINEXT-H~\cite{UNINEXT} & 81.8 & 77.7 & 85.8 & 85.4 & 89.0 & 86.4 \\
MITS~\cite{MITS} & 84.9 & 82.0 & 87.7 & 83.4 & 88.9 & 84.6 \\
OmniTracker-L~\cite{OmniTracker} & 71.0 & 66.8 & 75.2 & 83.4 & 86.7 & 82.3 \\
\rowcolor[rgb]{0.929,0.902,0.973}
Ours & \textbf{89.1} & \textbf{86.3} & \textbf{91.9} & \textbf{86.0} & \textbf{90.1} & \textbf{87.3}   \\
\bottomrule
\end{tabular}
    }
    \vspace{-4mm}
\end{table}

\textbf{Comparison with Unified Models.}
To demonstrate the superior performance of our unified tracking model, we conducted a comparative evaluation against other unified models. As shown in Table~\ref{app:table:sota_unifiedmodel}, our model achieves significantly better results on two classical benchmark datasets, highlighting its remarkable effectiveness and robustness.

\textbf{Qualitative Results.}
We first compare our method with representative SOTA models across three tracking tasks.
As shown in Fig.~\ref{fig:app:result_comp}, SAM 2++ produces more stable and accurate results across mask, box, and point tracking, demonstrating robustness and generalization to diverse scenarios.
We further visualize the multi-granularity tracking capability of SAM 2++ across multiple benchmarks in Fig.~\ref{fig:app:model_result}, where the same unified model handles different target state granularities with correct predictions.

\begin{table*}[t]
    \caption{Analysis of Mixed training strategy and Data Engine. \\ `Shared' and `Decoupled' denote that the module shares parameters or decouples parameters between different tasks, respectively.}
    \label{tab:ablation_data}
    \vspace{-1mm}
    \centering
    \resizebox{0.84\textwidth}{!}{
    \begin{tabular}{c|c|c|cccccc}
\toprule
Phase & Image Encoder & Memory & \multicolumn{1}{c}{LVOS $_{v1}^{val}$} & \multicolumn{1}{c}{TAG$_{VOS}^{test}$} & OTB100 & TNL2K & BADJA & TAG$_{PT}^{test}$ \\
\midrule
$\times$     & Shared & Shared      & 69.3          & 84.7           & 28.1        & 26.1       & 64.1  & 30.3    \\
$\times$     & Shared      & Decoupled      & 73.6          & 86.4           & 64.5  & 57.3            & 63.0            & 28.7         \\
+\textcircled{1}     & Freezed       & Decoupled      & 74.2          & 87.1           & 67.8        & 59.2       & 68.8            & 29.9         \\
+\textcircled{1}     & Decoupled      & Decoupled      & 75.4          & 86.8           & 66.9        & 58.5       & 72.1            & 32.7         \\
+\textcircled{1}     & Shared      & Decoupled      & 76.4          & 87.1           & 68.9  & 58.2            & 71.9  & 33.1         \\
\rowcolor[rgb]{0.929,0.902,0.973}
+\textcircled{1},\textcircled{2},\textcircled{3}   & Shared      & Decoupled      & 77.8          & 87.4           & 70.6  & 59.2            & 72.8            & 35.2 \\
\bottomrule
\end{tabular}

    }
    \vspace{-2mm}
\end{table*}

\begin{table*}[t]
  \caption{Analysis of training data and task mixtures on three tracking tasks. The table compares single-task and mixed-task training settings, showing how joint training and additional TAG data affect performance across mask, box, and point tracking.}
    \vspace{-1mm}
  \label{app:table:ablation_data}
  \centering
    \resizebox{0.97\textwidth}{!}
    {
    \begin{tabular}{c|c|c|ccc|ccc|ccc|c|c}
\toprule
\multicolumn{1}{c|}{\multirow{2}{*}{Type}} & \multicolumn{1}{c|}{\multirow{2}{*}{Mixture}} & \multicolumn{1}{c|}{\multirow{2}{*}{+Phase \textcircled{1}}} & \multicolumn{3}{c|}{MOSE} & \multicolumn{3}{c|}{GOT10K - val} & \multicolumn{3}{c|}{LaSOT} & \multicolumn{1}{c|}{\multirow{2}{*}{BADJA}} & \multicolumn{1}{c}{\multirow{2}{*}{TAG$_{PT}^{test}$}} \\ \cline{4-12}
\multicolumn{1}{c|}{}    & \multicolumn{1}{c|}{} & \multicolumn{1}{c|}{}   & $\mathcal{J}\&\mathcal{F}$ & $\mathcal{J}$ & $\mathcal{F}$      & AUC & $P_{Norm}$ & $P$  & AUC     & $P_{Norm}$ & $P$      & \multicolumn{1}{c|}{}     & \multicolumn{1}{c}{}   \\
\midrule
SAM 2  & -   & -     & 73.6    & 77.6   & 69.5   & 82.0      & 92.2 & 81.6     & 65.4    & 72.6    & 69.7   & $\times$ & $\times$    \\
\midrule
\#1    &     & \checkmark     & - & -      & -      & 86.3      & 94.7 & 87.3     & 68.7    & 75.4    & 73.5   & 66.2    & 78.6 \\
\#2    & \checkmark   & & 74.4    & 70.4   & 78.4   & 85.8      & 94.1 & 86.5     & 68.8    & 75.7    & 73.5   & 63.0    & 71.3 \\
\rowcolor[rgb]{0.929,0.902,0.973}
\#3    & \checkmark   & \checkmark     & 74.7    & 70.6   & 78.8   & 86.7      & 95.4 & 88.6     & 70.9    & 78.3    & 76.7   & 71.9    & 81.4   \\
\bottomrule
\end{tabular}
    }
    \vspace{-2mm}
\end{table*}

\subsection{Exploration Studies}

\textbf{Study on Mixed training strategy.}
To verify the effectiveness of task-adaptive memory during multi-task joint training, we compare the results of single-task training with different parameter settings during multi-task mixing training.
As shown in Table~\ref{tab:ablation_data}, when a single set of parameters is naively shared for multi-task joint training, the differences between tasks lead to a performance drop across all tasks. This indicates that the encoding and retrieval components of the memory module need to be decoupled for different tasks. In addition, when the image encoder is either frozen or similarly decoupled, the performance is inferior to a shared encoder. This suggests that the image encoder benefits from exposure to more data and is not adversely affected by task differences.

\textbf{Study on data engine.}
To validate the effectiveness of our data engine, we evaluated the performance when trained on different phases of our TAG dataset, as shown in Table~\ref{tab:ablation_data}. The results demonstrate that our proposed dataset enhances the performance on other datasets, indicating high diversity and generalizability.
After training with more data from phases \textcircled{2} and \textcircled{3}, the performance is further improved across all three tasks, demonstrating the effectiveness of the supplementary data provided by our data engine.

\textbf{Study on task mixture and training data.}
We compare the performance under different training settings for three tracking tasks in Table~\ref{app:table:ablation_data}.
For evaluating original SAM 2 on the single object task, we take the ground-truth bounding box from the first frame as a box prompt to predict the target mask, then predict the mask frame by frame, and finally extract the outer bounding box from each mask as the final box prediction.
After training on the public dataset and Phase \textcircled{1} of our Tracking-Any-Granularity dataset, the performance of our SAM 2++ model improves across all three tasks, demonstrating the advantages of our model design.
More importantly, when we further incorporate two additional tasks during training, the model's performance on both tasks surpasses that of training on a single task alone.
This illustrates two core motivations behind our proposed model:
(1) Although the granularity of the target states in the three tasks differs, they all can adopt the "matched memory" tracking paradigm. Thus, training on various tasks enhances the matching ability, which in turn improves the performance of all tracking tasks.
(2) As a generalized model supporting multiple tasks, SAM 2++ can be trained on large-scale datasets for multiple tasks, rather than being restricted to individual tasks.
Finally, under the task-mixed training setting, incorporating our proposed dataset further improves the model performance on both tasks. This improvement demonstrates that the diverse and comprehensive annotations included in our dataset provide valuable supervision signals for the model, enabling it to learn more robust and generalizable representations.

To make a fairer comparison, we added the results for the mask and bounding box tasks using separate public datasets for each task to Table~\ref{app:table:public_train} (point tracking is excluded because keypoint data is insufficient for individual training). Our model still outperforms the dedicated baselines, demonstrating that our model's performance advantage lies not only in its ability to utilize multi-task datasets but also in the superiority of its architectural design.

\begin{table}[t]
  \caption{Single-task comparison using public training datasets. We compare SAM 2++ trained separately for mask and box tracking with dedicated task-specific baselines under comparable public-data settings.}
  \label{app:table:public_train}
  \centering
    \resizebox{0.80\columnwidth}{!}
    {
    \begin{tabular}{l|ccc}
\toprule
Model & \multicolumn{3}{c}{LVOS$_{v2}^{val}$} \\
\midrule
OneVOS$_{\text{w/MOSE}}$    & 74.7  & 71.1  & 78.3 \\
\rowcolor[rgb]{0.929,0.902,0.973}
Ours (mask training only) & \textbf{81.6}   & \textbf{78.2}  & \textbf{85.1} \\
\midrule
\midrule
Model & \multicolumn{3}{c}{VastTrack} \\
\midrule
SUTrack$_{\text{224}}$ & 44.2 & 53.8 & 44.8 \\
\rowcolor[rgb]{0.929,0.902,0.973}
Ours (box training only) & \textbf{50.1} & \textbf{59.6} & \textbf{53.2} \\
\bottomrule
\end{tabular}

    }
    \vspace{-3mm}
\end{table}

\begin{table}[t]
    \caption{Analysis of the model setting on point tracking task.}
    \label{app:table:ablation_model_point}
    \vspace{-1mm}
    \centering
    \begin{tabular}{l|cc|c}
\toprule
Type & prompt & coordinates & BADJA   \\
\midrule
\#1 & Coord. & $argmax$(mask) & 65.6  \\
\#2 & Coord. \& Gauss. Mask & MLP    & 64.8 \\
\rowcolor[rgb]{0.929,0.902,0.973}
\textbf{\#3} & Coord. \& Gauss. Mask & $argmax$(mask) & 66.2 \\
\bottomrule
\end{tabular}

    \vspace{-3mm}
\end{table}

\begin{table}[t]
    \caption{Hyperparameters and details of SAM 2++ training in three tasks.}
    \label{tab:app:model_cost_details}
    \vspace{-1mm}
    \centering
    \resizebox{1.0\columnwidth}{!}{
        \begin{tabular}{c|ccc}
\toprule
\textbf{Modules} &  \textbf{GFlops} &  \textbf{Main Param} & \textbf{\textcolor{gray}{Lora Param}} \\
\midrule
 Image Encoder &  264.4     &  69.1     &  22.25       \\
 Memory Encoder   &  5.0    &  1.4 $\times$ 3     & -       \\
 Mask Decoder  &  53.4   &  9.9  & -      \\
 Memory Attention &  27.4   &  5.9     &  4.3  $\times$ 3     \\
 \textbf{Total}   &  350.2     &  89.1     & 35.1  \\
 \bottomrule  
\end{tabular}
    }
    \vspace{-3mm}
\end{table}

\begin{figure*}[t]
    \centering
    \includegraphics[width=\linewidth]{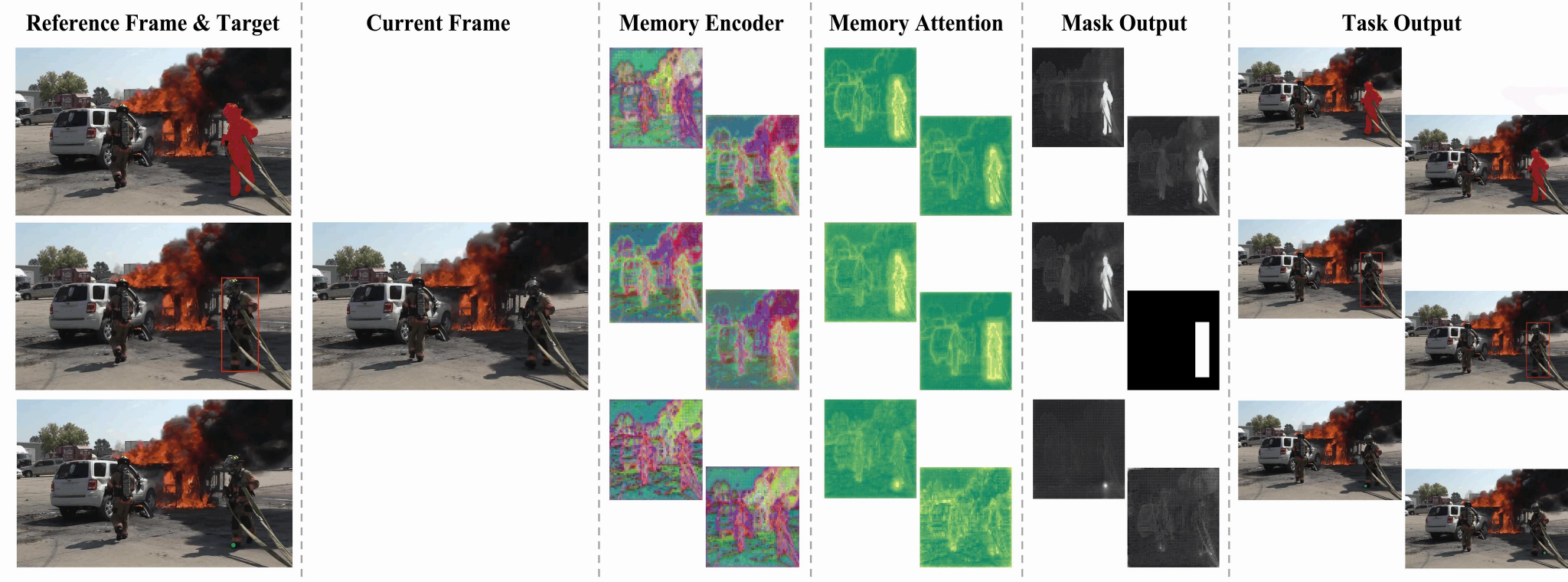}
    \vspace{-6mm}
    \caption{Visualization of memory design at different components and granularities.
    In the visualization of each component, the left side represents our task-adaptive memory mechanism, and the right side represents full parameter sharing.
    }
    \label{fig:visualization}
    \vspace{-4mm}
\end{figure*}

\textbf{Study on model setting of point tracking task.}
We compare the performance of point tracking under different model settings in Table~\ref{app:table:ablation_model_point}.
First, performance declines when the Gaussian mask prompt for the point tracking task is removed, indicating that incorporating the Gaussian mask effectively assists the mask output of the Decoder, and demonstrating the effectiveness of our proposed task-specific prompt.
Second, we compare two approaches for obtaining point coordinates: applying argmax to the mask output \textit{v.s.} adding an MLP to predict the coordinates directly.
The results show that the argmax operation yields better performance, suggesting that argmax is an effective method for point prediction, supervises the mask output as a memory source, and better represents the target state at the point granularity.
In contrast, the additional MLP requires adaptation to the original model and struggles to supervise the mask output effectively.

\subsection{Efficiency Analysis}

To present a comprehensive view of the model’s computational complexity and parameter overhead, we provide a detailed breakdown of the computational cost for each module in Table~\ref{tab:app:model_cost_details}, including GFLOPs, the number of parameters, and the \textcolor{gray}{LoRA parameters} introduced during training. We would like to kindly note that \textbf{the LoRA parameters exist only during training} and are merged into the main model weights at inference time. Therefore, \textbf{they do not introduce additional parameters or computational overhead during inference}. In addition, although the model contains multiple Memory Encoders designed for different granularities, only the branch corresponding to the current granularity is activated at inference, while the others remain inactive (excluded from computation), ensuring that no extra inference cost is incurred. In summary, the inherent inconsistency of memory is a problem that cannot be solved by a fully parameter-sharing design, while our task-adaptive memory mechanism has done its best to reduce decoupling and overhead to achieve unification.

\subsection{Visualization}
To further illustrate the varying requirements for memory representation of targets at different granularities, we visualize the memory-related outputs for the three tasks under both task-adaptive memory mechanism (left) and full parameter sharing (right), as shown in Fig.~\ref{fig:visualization}.
Firstly, we observe that even under different training settings, the memory features for the same task remain highly similar, indicating that different granularities have distinct memory requirements.
Secondly, under the memory-related decoupled training setting with our task-adaptive memory mechanism, both memory attention and mask output align more closely with the task outputs compared to full parameter sharing, highlighting the necessity of the decoupled design.
Finally, we find that under the full parameter sharing setting, the mask output for point tracking does not exhibit a Gaussian pattern, leading to incorrect predictions, while the training method with a decoupled design does not exhibit this error. This demonstrates that the decoupled design effectively preserves the specific needs of different tasks.

\section{Limitations, Impacts, and Future}
\label{app:limitation}

As a foundational model, SAM 2++ demonstrates strong performance in video tracking tasks across all three granularities, setting a new and powerful benchmark in the field of general video tracking.
As an annotation tool, SAM 2++ supports tracking multi-granularity, which greatly reduces the time and cost required to switch trackers between different application scenarios.
Furthermore, its ability to automatically generate annotations at multiple granularities provides an efficient and accurate tool platform for a wide range of research fields.

However, the model still has some limitations.
First, the current version does not yet support language- and audio-based references. Addressing this limitation requires integrating corresponding feature extractors into the Prompt Encoder to accommodate more types of reference states, as well as introducing relevant datasets for training.
Second, in our task-specific memory, some parameters of the memory-related modules are decoupled for different tasks. Although this mechanism only adds a minimal number of parameters, these parameters are supervised by a single task and cannot benefit from multi-task learning as the majority of shared parameters do. To address the issues caused by decoupled parameters, one approach is to employ an adapter that unifies memory across different granularities, another is to fuse the decoupled parameters and dynamically adjust their scaling according to the specific task.
Additionally, SAM 2++ still faces challenges in accurately tracking objects under severe occlusion, fast motion, and the presence of similar distractors. To further enhance model performance in these difficult scenarios, introducing motion modeling and specialized memory designs could be effective solutions.

\section{Conclusion}
We present SAM 2++, a unified framework for tracking targets at any granularity.
Instead of treating mask, box, and point tracking as separate problems, SAM 2++ unifies them through task-specific prompts, a Unified Decoder, and a task-adaptive memory mechanism that represents target states across granularities.
To provide unified training and testing data, we also introduce the Tracking-Any-Granularity dataset, built through a data engine that combines manual annotation with model-assisted completion, providing aligned three annotations in real-world videos.
Experiments across video object segmentation, single object tracking, and online point tracking show that SAM 2++ achieves strong performance at all three granularities.
We hope SAM 2++ can serve as a useful baseline for general tracking and inspire future research built around more flexible target representations and unified models.

\section{Acknowledgements}
This work is supported by the National Key R\&D Program of China (No. 2022ZD0160900), the Basic Research Program of Jiangsu (No. BK20250009), the Fundamental and Interdisciplinary Disciplines Breakthrough Plan of the Ministry of Education of China (No. JYB2025XDXM118), and the Collaborative Innovation Center of Novel Software Technology and Industrialization.

\ifCLASSOPTIONcaptionsoff
  \newpage
\fi

{\small
\bibliographystyle{ieee_fullname}
\bibliography{egbib}
}


\begin{IEEEbiography}[{\includegraphics[width=1in,height=1.25in,clip,keepaspectratio]{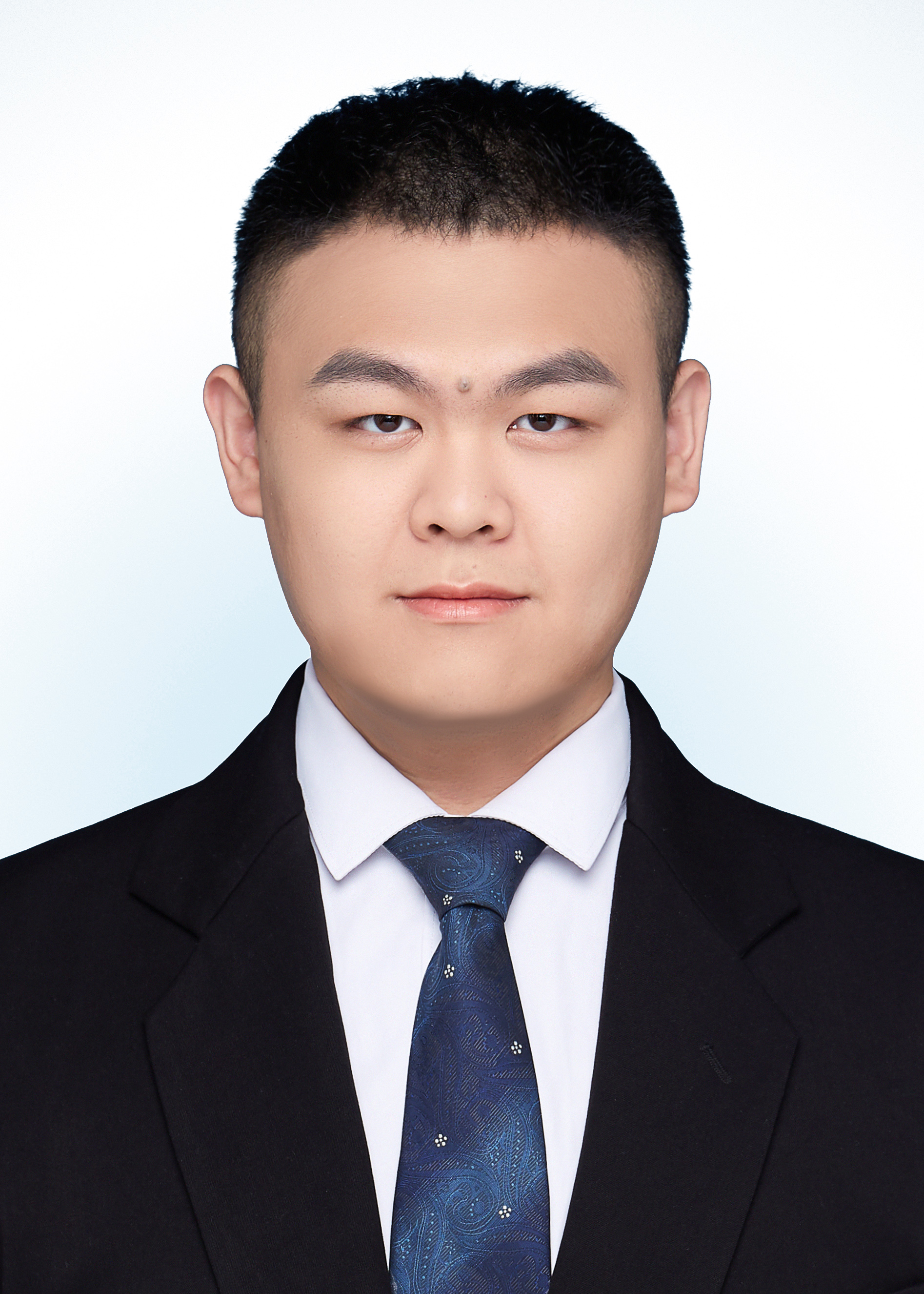}}]
 {Jiaming Zhang} received the BSc degree from Beijing Institute of Technology, Beijing, China, in 2021. He is currently working toward the PhD degree with the Department of Computer Science and Technology, Nanjing University from 2021.
 His research interests include computer vision and deep learning, with a focus on video segmentation and video understanding.
\end{IEEEbiography}

\begin{IEEEbiography}[{\includegraphics[width=1in,height=1.25in,clip,keepaspectratio]{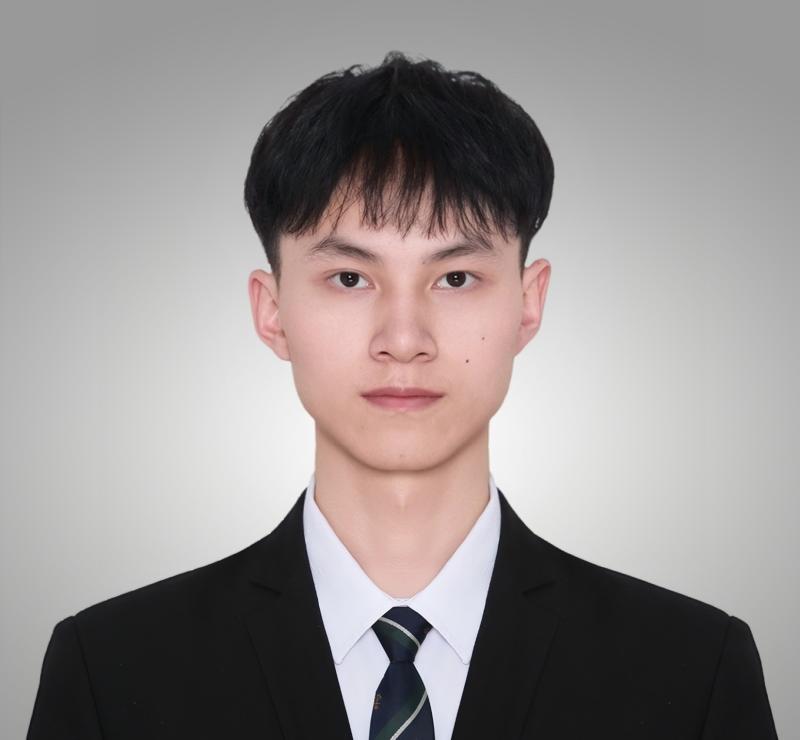}}]
 {Cheng Liang} received the BSc degree from Harbin Institute of Technology, Harbin, China, in 2023. He is currently working toward the MS degree with Nanjing University.
 His research interests include video generation, video analysis, and video object tracking.
\end{IEEEbiography}

\begin{IEEEbiography}[{\includegraphics[width=1in,height=1.25in,clip,keepaspectratio]{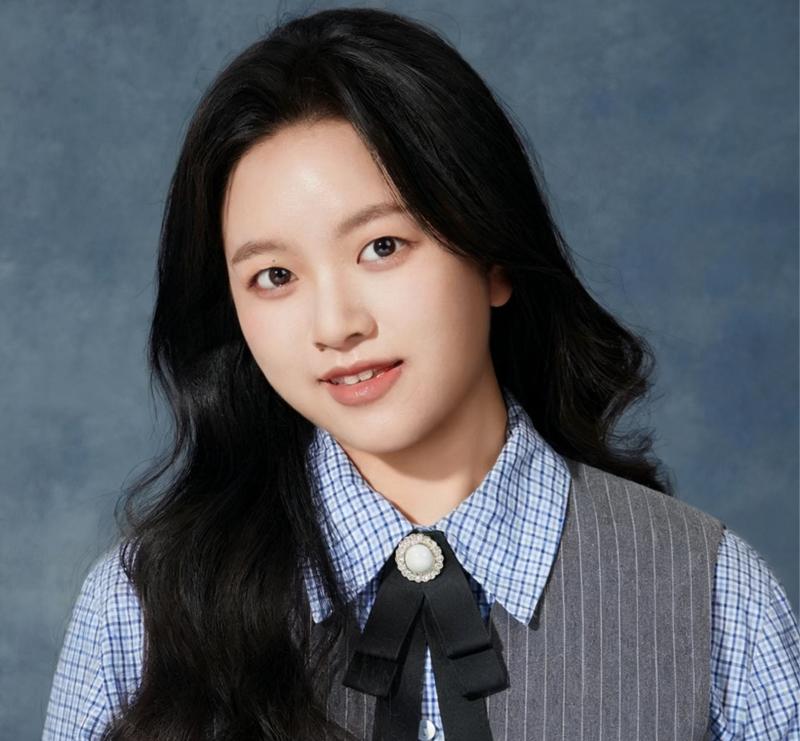}}]
 {Yichun Yang} received the BSc degree from Nanjing University, Nanjing, China, in 2022. He is currently working toward the MS degree with Nanjing University.
 His research interests include computer vision and deep learning.
\end{IEEEbiography}

\begin{IEEEbiography}[{\includegraphics[width=1in,height=1.25in,clip,keepaspectratio]{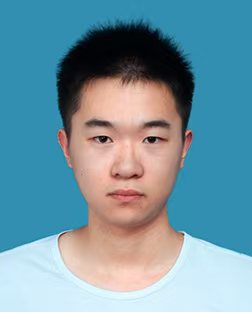}}]
 {Chenkai Zeng} received the BSc degree from Nanjing University, Nanjing, China, in 2022, and the MS degree from the School of Computer Science and Technology, Nanjing University, in 2025.
 His research interests include computer vision and deep learning.
\end{IEEEbiography}

\begin{IEEEbiography}[{\includegraphics[width=1in,height=1.25in,clip,keepaspectratio]{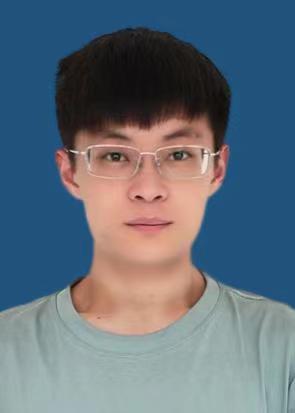}}]
 {Yutao Cui} received the B.Sc. and M.S. degree from Beijing Institute of Technology, Beijing, China, in 2017 and 2019 respectively, and the PhD degree from the Department of Computer Science and Technology, Nanjing University, Nanjing, China in 2025. His current research interests include visual object tracking and video object segmentation.
\end{IEEEbiography}

\begin{IEEEbiography}[{\includegraphics[width=1in,height=1.25in,clip,keepaspectratio]{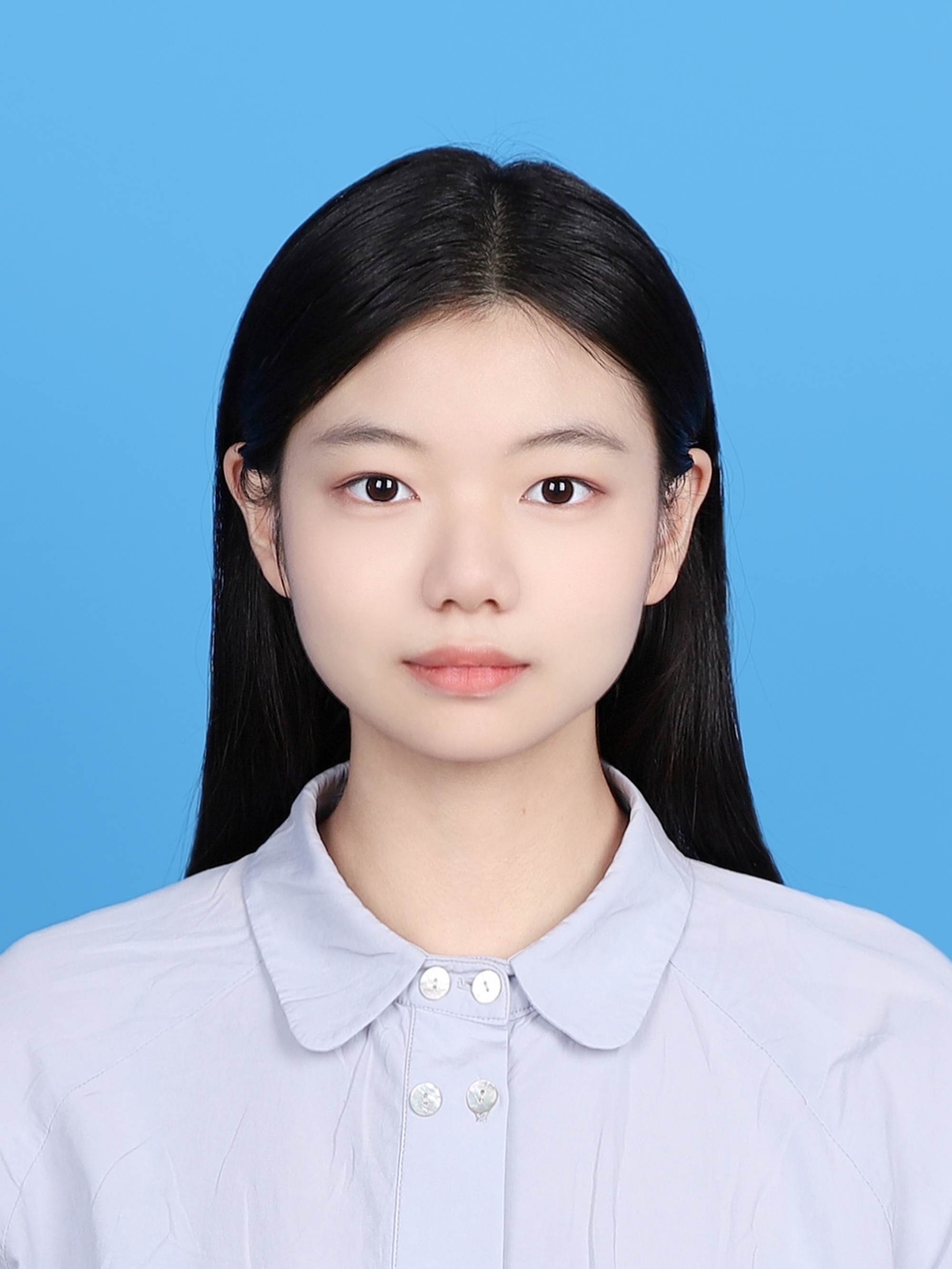}}]
 {Xinwen Zhang} received the BSc degree from the School of Artificial Intelligence, Nanjing University, Nanjing, China, in 2024. Xinwen Zhang is currently working toward the MS degree with the School of Artificial Intelligence, Nanjing University.
 Xinwen Zhang's research interests include computer vision and deep learning.
\end{IEEEbiography}

\begin{IEEEbiography}[{\includegraphics[width=1in,height=1.25in,clip,keepaspectratio]{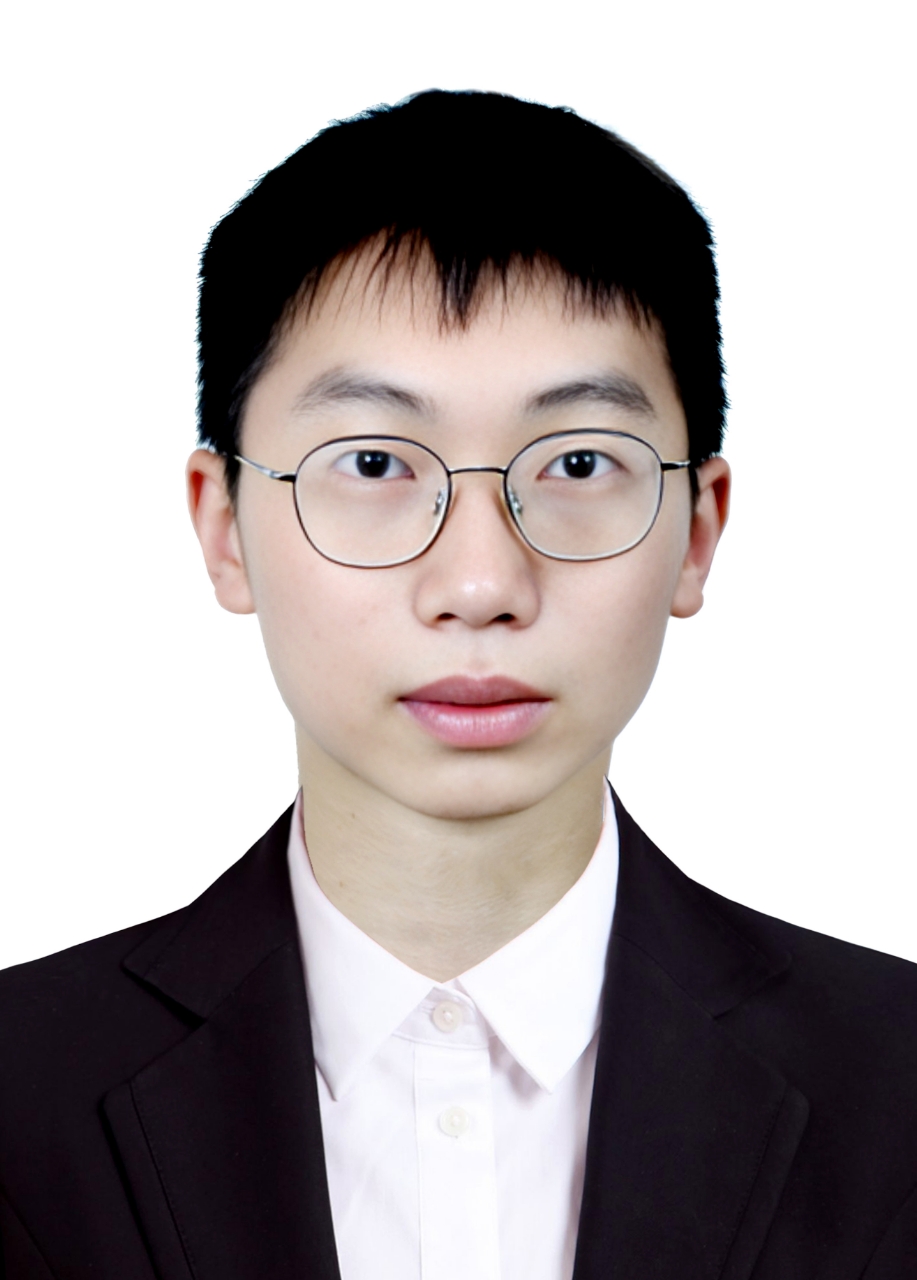}}]
 {Xin Zhou} received the BSc degree from Nanjing University, Nanjing, China, in 2024. Xin Zhou is currently working toward the MS degree with Nanjing University.
 Xin Zhou's research interests include computer vision and deep learning.
\end{IEEEbiography}

\begin{IEEEbiography}[{\includegraphics[width=1in,height=1.25in,clip,keepaspectratio]{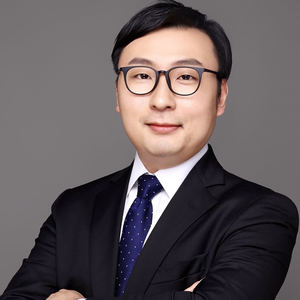}}]
 {Kai Ma} received the PhD degree in electrical and computer engineering from the University of Illinois at Chicago, Chicago, USA, in 2013. From 2014 to 2018, he was a staff research scientist with Siemens Corporate Research, Princeton, USA. He is currently a principal scientist with Tencent, Shenzhen, China.
 His research interests include deep learning, self-supervised learning, large model pretraining, and AIGC.
\end{IEEEbiography}

\begin{IEEEbiography}[{\includegraphics[width=1in,height=1.25in,clip,keepaspectratio]{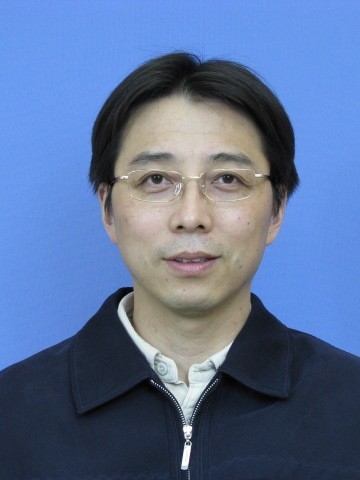}}]
 {Gangshan Wu} (Member, IEEE) received the BSc, MS, and PhD degrees from the Department of Computer Science and Technology, Nanjing University, Nanjing, China, in 1988, 1991, and 2000 respectively. He is currently a professor with the Department of Computer Science and Technology, Nanjing University. His current research interests include computer vision, multimedia content analysis, multimedia information retrieval, digital museum, and large-scale volumetric data processing.
\end{IEEEbiography}

\begin{IEEEbiography}[{\includegraphics[width=1in,height=1.25in,clip,keepaspectratio]{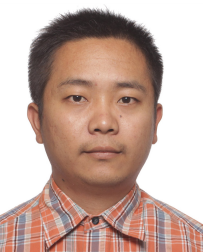}}]
 {Limin Wang} (Senior Member, IEEE) received the BSc degree from Nanjing University, Nanjing, China, in 2011, and the PhD degree from the Chinese University of Hong Kong, Hong Kong, in 2015. From 2015 to 2018, he was a post-doctoral researcher with Computer Vision Laboratory, ETH Zurich. He is
currently a professor with the School of Computer Science, Nanjing University. His research interests include computer vision and deep learning. He has
served as area chair for NeurIPS, CVPR, ICCV, and is an associate editor of the \emph{IEEE Transactions on Pattern Analysis and Machine Intelligence} and the \emph{International Journal of Computer Vision}.
\end{IEEEbiography}

\end{document}